\documentclass[preprint,10pt,numbers]{elsarticle}
\usepackage{setspace}
\usepackage{times}
\usepackage{xurl}
\usepackage{hyperref}
\usepackage{graphicx}
\usepackage{booktabs}
\usepackage{multirow}
\usepackage{wrapfig}
\usepackage[ruled,vlined]{algorithm2e}
\usepackage{amsmath,amssymb}
\usepackage{bbm}
\usepackage[table,dvipsnames,svgnames,x11names]{xcolor}
\usepackage{lineno}

\journal{}

\begin{document}

\begin{frontmatter}

\title{Rapid-Deployment Crack Measurement Based on SAM3 Semantic-Edge Response Decoding}

\author[2]{Zhanping Song}
\ead{songzhpyt@xauat.edu.cn}
\affiliation[2]{organization={Xi'an University of Architecture and Technology},
            department={School of Civil Engineering,},
            city={Xi'an},
            postcode={710055}, 
            state={Shaanxi},
            country={China}}
\author[1]{Shipeng Liu\corref{cor1}}
\ead{lsp@xauat.edu.cn}
\affiliation[1]{organization={Xi'an University of Architecture and Technology},
            department={School of mechanical and electrical engineering,},
            city={Xi'an},
            postcode={710055}, 
            state={Shaanxi},
            country={China}}
\author[3]{Liang Zhao\corref{cor1}}
\ead{zhaoliang@xauat.edu.cn}
\affiliation[3]{organization={Xi'an University of Architecture and Technology},
            department={School of Artificial Intelligence and Robotics,},
            city={Xi'an},
            postcode={710055}, 
            state={Shaanxi},
            country={China}}
\author[4]{Dengfeng Chen}
\ead{chdengf@xauat.edu.cn}
\affiliation[4]{organization={Xi'an University of Architecture and Technology},
            department={School of Building Equipment Science and Engineering,},
            city={Xi'an},
            postcode={710055}, 
            state={Shaanxi},
            country={China}}
\cortext[cor1]{Corresponding author.}

\begin{abstract}
Reliable crack measurement is essential for infrastructure condition assessment, yet existing image-based approaches typically depend on pixel-wise annotations, task-specific segmentation training, and mask-based geometric measurement, making cross-scene deployment costly and sensitive to segmentation errors. We identify an output-interface mismatch in SAM3: its prompt-conditioned semantic response preserves crack evidence that is often suppressed or spatially distorted in the final candidate masks. Across six public crack datasets, the internal response achieves 82.66\% average crack-pixel recall, compared with 74.66\% for the retained SAM3 proposals, with an average mismatch ratio of 8.73\%. Based on this observation, we propose \textbf{S}emantic-\textbf{E}dge \textbf{R}esponse \textbf{D}ecoding (\textbf{SERD}) to calibrate the semantic response using a fixed Sobel structural field, and further develop \textbf{SERD-DQ}, a training-free framework that directly estimates crack centerline and transverse geometry from the continuous decoded response without generating an intermediate predicted mask. Experiments verify both segmentation fidelity and direct geometric measurement against manually established pixel-level references. Compared with native SAM3 mask-based quantification, SERD-DQ reduces width MAE from 6.072 to 5.547 pixels, length relative error from 22.245\% to 17.355\%, and area relative error from 33.921\% to 26.228\%, while achieving a latent geometry recovery rate of 0.394. The results indicate that continuous semantic-edge responses provide a more reliable interface for training-free crack quantification than conventional mask-mediated measurement.
\end{abstract}


\end{frontmatter}


\section{Introduction}
\label{sec1:intro}

Crack measurement is a fundamental component of structural inspection because maintenance decisions depend on quantitative defect information rather than visual detection alone. Crack length characterizes spatial extension and propagation, crack width is closely related to severity assessment, and crack area provides a compact description of the affected region. In tunnels, bridges, pavements, dams, buildings, and other large-scale structures, these quantities are traditionally obtained through on-site visual inspection, rulers, crack-width gauges, microscopes, or image-assisted manual measurement~\citep{konig2022whatscracking,xiang2023dtrcnet}. Although such procedures can provide reliable local observations, they are labor intensive, difficult to scale to large image collections, and strongly dependent on acquisition conditions and operator effort.

Computer vision has substantially automated this process, but most existing methods still follow a supervised \emph{segment-then-measure} pipeline. Inspection images must first be collected and densely annotated, a crack segmentation model is then trained for the target task, and geometric quantities are finally computed from the predicted binary mask by skeletonization, boundary extraction, distance transforms, or related operators. DeepCrack~\citep{liu2019deepcrack}, CrackFormer~\citep{liu2021crackformer}, DTrCNet~\citep{xiang2023dtrcnet}, SCSegamba~\citep{liu2025scsegamba}, and MixerCSeg~\citep{zhao2026mixercseg} demonstrate the strong segmentation accuracy achievable under this paradigm. However, its deployment cost remains substantial. Pixel-level annotation is expensive, task-specific models can be sensitive to changes in material, illumination, scale, contamination, and crack morphology, and each new inspection scene may require additional data preparation, training, and validation. More importantly, measurement inherits the errors of the segmentation interface: missed branches shorten the measured crack, over-segmented regions inflate area and width, and even a small boundary displacement can produce a large relative width error for thin cracks.

Promptable vision foundation models provide a route to reduce this redeployment cost. Models in the SAM family transfer general visual priors without crack-specific retraining~\citep{kirillov2023sam,ravi2024sam2}, and SAM3 further supports text-conditioned concept segmentation~\citep{carion2025sam3}. With the fixed prompt \texttt{crack}, the frozen model can expose crack-related evidence in unseen inspection images without target-domain optimization. However, directly measuring geometry from the native SAM3 proposals still retains the segment-then-measure bottleneck: weak branches may be omitted, crack-like textures may be over-segmented, and small boundary displacements can propagate into substantial width and area errors.

More broadly, this limitation raises an engineering-informatics question concerning how computational representations should be transformed into usable engineering knowledge. Engineering informatics has long emphasized that raw observations become valuable only when they are organized into representations that support interpretation, reasoning, and decision-making. For example, BIM-based knowledge graphs explicitly organize engineering entities and relations~\citep{teclaw2024bimkg}, infrastructure digital twins formalize heterogeneous system information through ontology- and graph-based models~\citep{li2024infrastructuredt}, and model--data integrated digital twins transform spatially discrete monitoring observations into full-field structural-state information~\citep{sun2024hybridtwin}. In vision-based inspection, however, this formalization is commonly mediated by task-specific outputs such as classes, bounding boxes, or segmentation masks, which are subsequently converted into geometric attributes, model updates, or condition indicators~\citep{wang2025asbuilt,chang2026tunnelcv}. These outputs are effective recognition interfaces, but they are compressed representations of the richer spatial evidence maintained by modern computational models. For measurement-oriented tasks, prematurely converting continuous evidence into a discrete task output may therefore discard topology, boundary variation, or geometric uncertainty that remains relevant to the engineering quantity of interest. This motivates a broader representation question: whether intermediate computational responses can themselves serve as knowledge-bearing interfaces from which structured and measurable engineering information is directly formalized.

Within crack measurement, we therefore examine whether the final mask is the most suitable SAM3 representation for formalizing the required engineering geometry. By statistically comparing the internal semantic response with the retained proposal masks, we observe a systematic \emph{output-interface mismatch}: a non-negligible portion of ground-truth crack evidence remains active in the prompt-conditioned semantic response but is absent from the final SAM3 proposals. Across the six crack datasets used in this study, the internal response reaches an average crack-pixel recall of 82.66\%, compared with 74.66\% for the final proposals, and the mean mismatch ratio is 8.73\%. This suggests that the continuous semantic response is not merely an intermediate activation but a more suitable representation for thin, elongated, and locally ambiguous crack structures.

Based on this observation, we propose \textbf{S}emantic-\textbf{E}dge \textbf{R}esponse \textbf{D}ecoding (\textbf{SERD}) and its direct-quantification formulation, \textbf{SERD-DQ}. SERD extracts the language-conditioned SAM3 semantic response and calibrates it with a fixed Sobel structural field, producing a continuous semantic-edge response that preserves crack semantics while strengthening structurally plausible boundaries. SERD-DQ then avoids the binary-mask bottleneck entirely: it traces crack centerlines and local cross-sections directly in the decoded response domain and obtains crack length, mean width, maximum width, and area from this continuous geometric representation.

Evaluation follows the same representation-to-measurement logic. SERD-Seg is introduced only as a temporary binary readout of the decoded response, allowing established region, boundary, and topology metrics to test whether the response preserves crack geometry. The final task is evaluated separately with manual pixel-domain references generated by a dedicated Crack Measurement Workbench on randomly sampled public images. This two-stage protocol distinguishes response fidelity from measurement accuracy and permits mask-based baselines and SERD-DQ to be compared against identical reference geometry.

The main contributions are summarized as follows:
\begin{itemize}
\item We identify and quantify an output-interface mismatch in SAM3, showing that its internal prompt-conditioned semantic response preserves crack evidence that is often lost or spatially distorted in the final masks.
\item We propose SERD-DQ, a training-free and mask-free crack measurement framework that calibrates SAM3 semantic responses with fixed Sobel edge evidence and directly estimates crack length, mean width, maximum width, and area from response-domain geometry.
\item We establish a two-level validation protocol that evaluates both crack representation quality and direct geometric measurement, using segmentation metrics for region, boundary, and topology fidelity and manually established pixel-level references for quantitative assessment.
\end{itemize}

\section{Related Work}

\subsection{Computational Formalization of Engineering Knowledge}

A central objective of engineering informatics is to transform heterogeneous observations and computational outputs into structured representations that can support engineering interpretation and decision-making. Such representations need not correspond directly to the raw form of the acquired data. In BIM and digital-twin research, for example, Teclaw et al.~\citep{teclaw2024bimkg} federated domain-specific BIM models into a knowledge graph so that engineering entities and their cross-domain relations could be represented explicitly. Li et al.~\citep{li2024infrastructuredt} developed an ontology- and graph-based infrastructure digital-twin paradigm to organize heterogeneous infrastructure information into a knowledge-rich model, while Sun et al.~\citep{sun2024hybridtwin} integrated monitoring data, finite-element models, BIM, and IoT to convert spatially discrete observations into full-field structural-state information. These studies illustrate a recurring engineering-informatics principle: computational value depends not only on acquiring or predicting data, but also on selecting an information representation that preserves the relationships required by subsequent engineering reasoning.

A similar representation chain appears in vision-based engineering inspection, although the intermediate knowledge representation is usually a recognition output. Ye et al.~\citep{ye2024samdamage} used SAM-based instance segmentation to automate structural-damage detection and subsequently estimated masonry-crack dimensions. Wang et al.~\citep{wang2025asbuilt} combined fine-grained recognition using vision foundation models with object-aware Scan-vs-BIM matching to transform perceptual observations into informative as-built digital-twin updates. More directly related to structural assessment, Chang et al.~\citep{chang2026tunnelcv} extracted crack information from images and mapped the resulting geometry into finite-element model updates for shield-tunnel condition evaluation. In tunnel seepage assessment, Wei et al.~\citep{wei2026samseepage} similarly used segmentation as the basis for downstream quantification and severity visualization. These studies extend visual recognition toward engineering interpretation, but they largely retain a predicted class, object, or mask as the interface through which visual information is formalized.

The same interface assumption is evident in crack-analysis research even when the learning strategy changes. Encoder--decoder models directly learn pixel-wise crack regions from dense supervision~\citep{fan2020automaticcrack}, while weakly supervised and semi-supervised methods reduce annotation requirements without changing the final pixel-mask representation~\citep{inoue2023weakcrack,shamsabadi2024semisupervised}. These advances are important for reducing data-preparation cost, but they primarily optimize how the mask is obtained rather than whether the mask is the most suitable carrier of geometric information for subsequent engineering measurement.

This observation exposes a representation-level question that has received less attention than recognition accuracy itself. A task output is designed to satisfy a prediction objective, whereas an engineering-information representation must preserve the attributes required by downstream analysis. The two requirements are not necessarily identical. In particular, when engineering decisions depend on local geometry, connectivity, or continuously varying dimensions, a discrete recognition output may remove information that is weakly relevant to classification or segmentation but strongly relevant to measurement. Our work investigates this distinction by asking whether the final segmentation mask should remain a compulsory knowledge interface when richer spatial evidence is already available inside a pretrained computational model.

\subsection{Geometry-Preserving Representations Beyond Binary Masks}

Evidence from other visual-analysis domains suggests that the representation best suited to geometric reasoning can differ from the representation optimized for regional classification. This issue is particularly evident for thin and network-like structures. Shit et al.~\citep{shit2021cldice} showed across vessels, roads, and neurons that preserving centerline connectivity captures a property that conventional region-overlap objectives do not explicitly represent. In road-network extraction, Bastani et al.~\citep{bastani2018roadtracer} demonstrated that inferring a road graph directly can avoid connectivity errors introduced when a noisy pixel-wise segmentation must first be converted into a graph. For tubular structures, Sironi et al.~\citep{sironi2014centerline} represented centerline localization through a learned scale-space distance transform, allowing both center position and local scale to be recovered from a continuous regression field. More recently, Liao~\citep{liao2023tubular} jointly recovered segmentation masks and centerlines through a minimal-path formulation rather than determining one representation only as a post-processing consequence of the other.

Boundary- and topology-oriented segmentation studies provide complementary evidence that regional overlap alone does not fully characterize geometry. Boundary IoU was introduced specifically because conventional Mask IoU can overlook important contour errors~\citep{cheng2021boundaryiou}; PointRend and SegFix improve segmentation by explicitly revisiting uncertain or inaccurate boundary regions~\citep{kirillov2020pointrend,yuan2020segfix}. For structural cracks, TOPO-Loss further demonstrates that continuity must be modeled explicitly because disconnected predictions can remain problematic even when pixel-wise accuracy is otherwise competitive~\citep{pantoja2022topoloss}. Together, these works support separating regional recognition quality from the preservation of geometry required by downstream measurement.

Although these studies address different domains, they reveal a common computational insight: preserving task-relevant geometry may require an intermediate representation whose structure is closer to the downstream quantity of interest than a generic binary region is. Centerlines encode longitudinal organization, distance or scale fields retain continuous transverse information, and graphs explicitly preserve connectivity. Importantly, however, most such approaches construct or learn these geometry-aware representations specifically for the target task. They therefore do not answer whether useful geometry-bearing representations may already exist inside a large pretrained model and can be exploited without learning an additional task-specific predictor.

Crack measurement provides a stringent engineering instance of this broader representation problem. Regional support, longitudinal continuity, and transverse width are simultaneously required, while a thin structure can suffer substantial relative geometric error from only a small mask displacement. Rather than learning another geometry-aware network, we investigate whether the prompt-conditioned continuous response already present in a frozen foundation model can be reorganized into an explicit centerline and width representation. The objective is therefore not to replace binary segmentation in general, but to test when a continuous computational representation provides a better knowledge interface for quantitative engineering analysis.

\subsection{Foundation Models as Engineering Knowledge Interfaces}

Vision foundation models further change the representation question because rich transferable features can be obtained before engineering-task-specific optimization. The progression from CLIP's transferable image--text representation~\citep{radford2021clip} to dense language-guided segmentation methods such as CLIPSeg, SAN, and CAT-Seg~\citep{luddecke2022clipseg,xu2023san,cho2024catseg} illustrates how generic pretrained semantics can be reorganized into pixel-level spatial interfaces. SAM~\citep{kirillov2023sam}, SAM2~\citep{ravi2024sam2}, and SAM3~\citep{carion2025sam3} further extend promptable segmentation toward increasingly transferable visual and semantic interaction. In civil infrastructure, CrackSAM shows that parameter-efficient adaptation of SAM can substantially improve crack segmentation and cross-scene robustness~\citep{ge2024cracksam}; however, the adapted knowledge is still ultimately communicated through a segmentation mask.

Engineering studies have begun to exploit these capabilities beyond recognition alone. Ye et al.~\citep{ye2024samdamage} developed automated SAM-based structural-damage instance segmentation and coupled detected masonry cracks with dimensional estimation. Xu et al.~\citep{xu2025samdimension} used SAM-based segmentation together with connected-domain and local-unit analysis to quantify full cross-sectional dimensions of structural components. Wei et al.~\citep{wei2026samseepage} integrated a SAM encoder into a tunnel-seepage segmentation framework and subsequently quantified seepage characteristics, while Chang et al.~\citep{chang2026tunnelcv} transferred image-derived crack geometry into finite-element model updates for structural assessment. Collectively, these studies demonstrate that foundation models can reduce the gap between generic visual perception and engineering information extraction. Nevertheless, the final segmentation result remains the principal interface through which the visual model communicates with the subsequent engineering-analysis stage.

At the same time, general computer-vision research has shown that the internal representations of pretrained models contain spatial information that need not be limited to their original image-level or final-output interfaces. DenseCLIP~\citep{rao2022denseclip} converts pretrained image--text knowledge into pixel--text score maps for dense prediction, while MaskCLIP demonstrates that dense semantic labels can be extracted from CLIP with minimal modification and without conventional dense annotation~\citep{zhou2022maskclip}. More recent training-free approaches such as NACLIP~\citep{hajimiri2025naclip}, CorrCLIP~\citep{zhang2025corrclip}, and ReME~\citep{xuan2025reme} reorganize pretrained representations, patch relations, or reference features to recover spatially localized semantics without conventional task-specific training. These studies establish that latent foundation-model representations can support spatial reasoning beyond their original output formulation. Their endpoint, however, remains a segmentation map.

\begin{table}[t]
\centering
\small
\caption{Representative computational representations used to formalize engineering or geometry-related information. The comparison focuses on the representation transferred from perception or sensing to downstream engineering analysis, rather than on prediction accuracy.}
\label{tab:knowledge_interface}
\setlength{\tabcolsep}{5pt}
\renewcommand{\arraystretch}{1.18}
\resizebox{\textwidth}{!}{%
\begin{tabular}{p{3.1cm}p{3.0cm}p{5.0cm}p{5.3cm}}
\toprule
\textbf{Representative work} &
\textbf{Engineering context} &
\textbf{Representation passed to downstream analysis} &
\textbf{Formalized engineering / geometric knowledge} \\
\midrule

Teclaw et al.~\citep{teclaw2024bimkg} &
BIM integration &
Entity--relation knowledge graph &
Cross-domain building entities and relations \\

Sun et al.~\citep{sun2024hybridtwin} &
Structural monitoring &
Model--data integrated digital-twin state &
Full-field structural-state information \\

Ye et al.~\citep{ye2024samdamage} &
Structural damage inspection &
SAM-derived damage instances and masks &
Damage localization and crack dimensions \\

Xu et al.~\citep{xu2025samdimension} &
Structural component measurement &
SAM segmentation with connected-region analysis &
Cross-sectional component dimensions \\

Chang et al.~\citep{chang2026tunnelcv} &
Tunnel condition assessment &
Image-derived crack masks and geometry &
Finite-element model updates and structural condition \\

Bastani et al.~\citep{bastani2018roadtracer} &
Road-network extraction &
Explicitly constructed graph &
Road connectivity and network topology \\

Sironi et al.~\citep{sironi2014centerline} &
Tubular-structure analysis &
Continuous scale-space distance field &
Centerline location and local structural scale \\

\rowcolor{gray!15}
SERD-DQ (ours) &
Crack measurement &
Continuous semantic-edge response and response-domain geometry &
Centerline, width profile, length, width, and area \\

\bottomrule
\end{tabular}}
\end{table}

\subsection{Comparative Perspective on Representation-to-Engineering Formalization}

The studies reviewed above suggest that an important distinction lies not only in the computational method being used, but also in the representation through which computational results are transferred to downstream engineering analysis. Engineering-informatics research has formalized observations through knowledge graphs, digital-twin states, recognized engineering entities, and mask-derived geometry, whereas geometry-oriented studies from other domains have employed graphs, centerlines, and continuous fields to preserve connectivity or continuously varying spatial attributes. Table~\ref{tab:knowledge_interface} summarizes representative examples from this perspective. Rather than comparing predictive performance, the table focuses on what information representation is retained at the interface between perception or sensing and subsequent engineering reasoning.

The comparison reveals two related trends. First, engineering-information formalization increasingly involves representations that are more structured than raw sensing observations, such as entity relations, digital-twin states, and explicit geometric attributes. Second, studies concerned with geometry and topology show that the most useful downstream representation does not necessarily coincide with a conventional pixel-wise region. Graphs can directly encode connectivity, while centerline and continuous-field representations can retain spatial structure that may be weakened when information is reduced to a binary region.

The missing connection between these research streams is how latent representations from pretrained foundation models can be directly formalized into engineering quantities. Existing engineering applications of foundation models predominantly convert pretrained knowledge into a task-specific output, most commonly a segmentation mask, and subsequently derive engineering information from that output. In contrast, representation-oriented vision studies demonstrate that useful spatial evidence can remain available before the final prediction interface, but generally reorganize this evidence to obtain another segmentation result. It therefore remains unclear whether such internal evidence can itself serve as an engineering-information representation from which measurable quantities are directly constructed.

SERD-DQ addresses this gap by treating the prompt-conditioned SAM3 semantic response as the information-bearing representation rather than merely as an intermediate activation used to generate a mask. After structural calibration, the continuous response is organized into a crack centerline and transverse width profile and is subsequently used to estimate length, mean width, maximum width, and area. The distinction from previous work therefore lies not simply in replacing one segmentation method with another, but in changing the representation interface through which pretrained visual knowledge is formalized for engineering measurement.

\begin{figure}[t]
    \centering
    \includegraphics[width=1\linewidth]{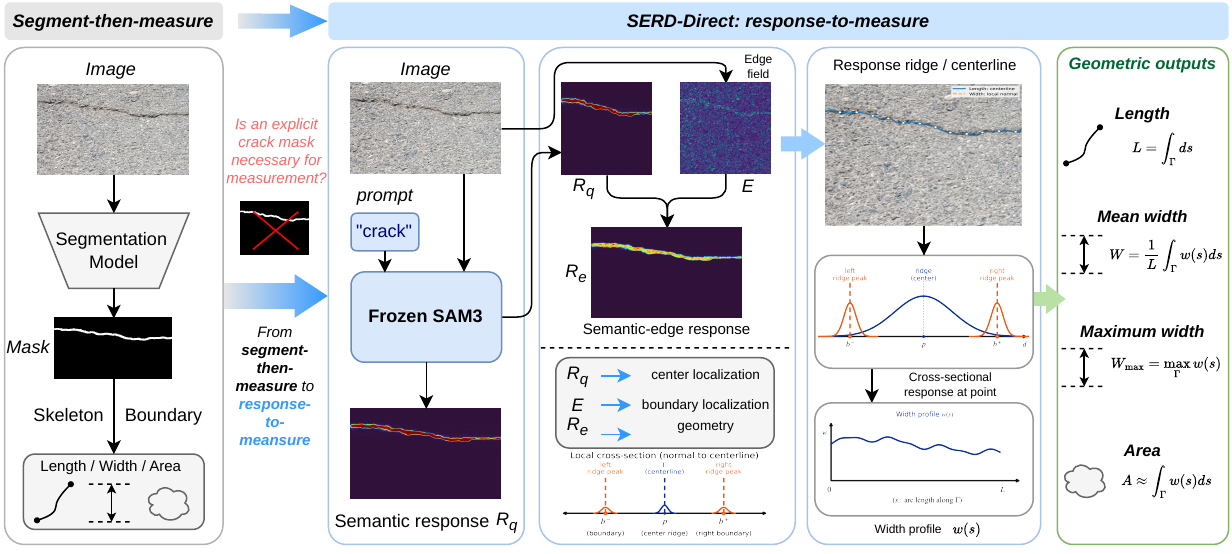}
    \caption{Overall architecture of the proposed SERD-DQ framework. Conventional crack measurement typically follows a segment-then-measure pipeline, where an explicit binary mask is first generated and subsequently processed to extract the crack skeleton, boundary, and geometric quantities. In contrast, SERD-DQ adopts a response-to-measure paradigm. Given an image and the text prompt $\texttt{crack}$, frozen SAM3 produces a prompt-conditioned semantic response $R_q$, which is further calibrated with the Sobel edge field $E$ to obtain the semantic-edge response $R_e$. The decoded response is then used to trace the crack centerline and estimate cross-sectional boundaries along the local normal direction, yielding a continuous width profile $w(s)$. Crack length, mean width, maximum width, and area are directly quantified from the centerline and width profile without generating an intermediate predicted crack mask.}
    \label{fig:framework}
\end{figure}

\section{Method}
\label{sec:method}

\subsection{Overall Framework and Problem Formulation}

Given an inspection image $I\in\mathbb{R}^{H\times W\times3}$ and the fixed prompt $q=\text{\texttt{crack}}$, a conventional mask-based measurement pipeline can be written as
\begin{equation}
(I,q)\rightarrow M\rightarrow \mathcal{G}(M)\rightarrow \{L,\overline W,W_{\max},A\},
\end{equation}
where $M$ is a predicted crack mask and $\mathcal{G}$ is a deterministic geometry extractor. This formulation makes segmentation an obligatory intermediate interface, so missed branches, false regions, and local boundary displacement are directly inherited by the measured geometry.

SERD-DQ instead treats the internal SAM3 response as the measurement representation. SAM3 remains completely frozen and exposes a prompt-conditioned semantic response $R_q$. A fixed structural field $E$ is used to decode a semantic-edge response
\begin{equation}
R_e=\mathcal{D}(R_q,E),
\end{equation}
where $\mathcal{D}$ denotes Semantic-Edge Response Decoding. SERD-DQ then maps the continuous response directly to centerline and transverse geometry,
\begin{equation}
(I,q)\rightarrow R_q\rightarrow R_e\rightarrow \{\Gamma,w(s)\}\rightarrow \{L,\overline W,W_{\max},A\}.
\end{equation}
No predicted binary mask is generated or consumed in this measurement path.

\begin{figure}[t]
    \centering
    \includegraphics[width=1\linewidth]{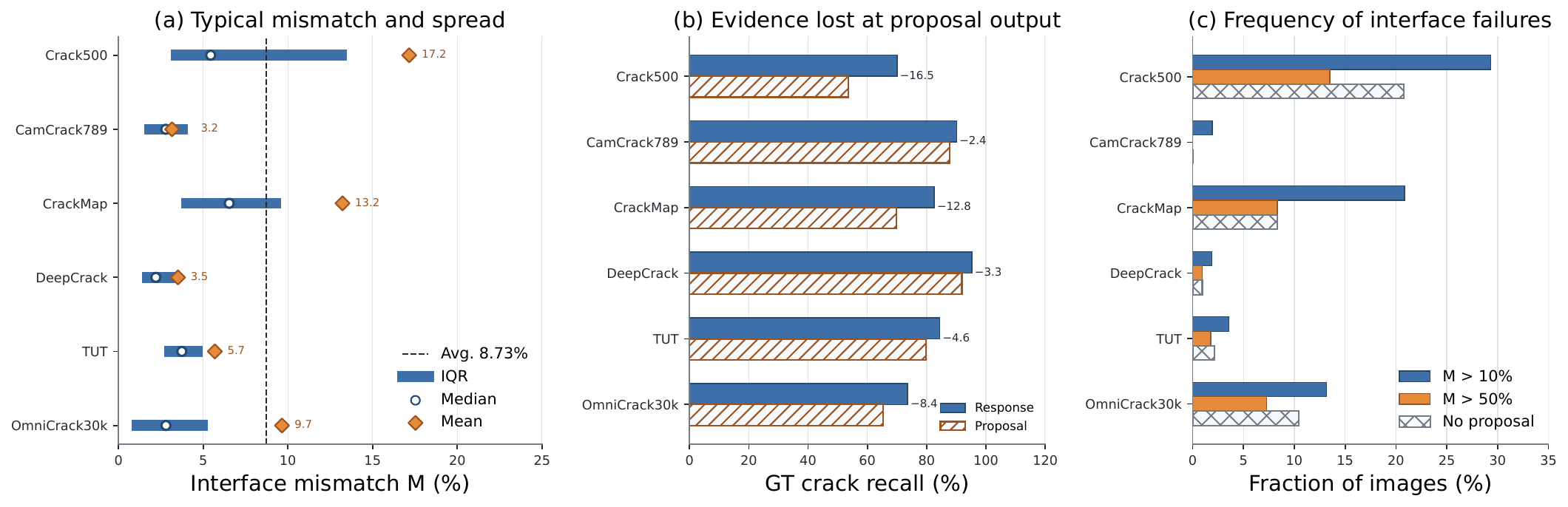}
    \caption{Mismatch between the internal SAM3 semantic response and the final candidate-mask interface on six public crack datasets. (a) Distribution of the crack-evidence mismatch ratio $\mathcal{M}_i$. (b) Crack-pixel recall of the internal response and retained SAM3 proposals. (c) Proportions of images with substantial mismatch or no retained proposal. The analysis shows that crack evidence available inside the frozen model can be lost or spatially altered before the final mask interface.}
    \label{fig:interface_mismatch}
\end{figure}

\subsection{Prompt-Conditioned Semantic Response Extraction}
\label{sec:response_extract}

SAM3 accepts the concept prompt \texttt{crack} and produces an internal language-conditioned semantic response in addition to its final region proposals. Let the semantic logit map exposed by the segmentation head be $Z_q$. After sigmoid activation, restoration to the evaluation resolution, and per-image normalization, the response is written as
\begin{equation}
R_q=\operatorname{Norm}\!\left(\operatorname{Resize}(\sigma(Z_q))\right).
\end{equation}
We treat $R_q$ as a relative crack-evidence field rather than a calibrated foreground probability. Unlike the final proposal masks, $R_q$ retains continuous response magnitude and spatial continuity before proposal confidence filtering, mask thresholding, and proposal merging. This property is important for direct measurement because longitudinal crack evidence and local transverse response changes can be exploited without first committing every pixel to a binary class. Code-level extraction from the official SAM3 implementation is provided in~\ref{app:response_extract}.

\paragraph{\textbf{Response--mask interface mismatch}}
The choice of $R_q$ as the primary crack representation is supported by an empirical analysis of the SAM3 output interface. For image $i$, let $A_i$ denote the active semantic-response set, $B_i$ the union of retained native SAM3 candidate masks, and $P_i$ the ground-truth crack set. We define
\begin{equation}
\mathcal{M}_i=\frac{|(A_i\cap P_i)\setminus B_i|}{|P_i|},
\end{equation}
which measures the fraction of annotated crack pixels that are activated by the internal semantic response but disappear from the final candidate-mask interface. Across 4,634 crack-bearing images from six public datasets, the equal-weight mean mismatch ratio is 8.73\%. The internal semantic response reaches 82.66\% average crack-pixel recall, compared with 74.66\% for the retained SAM3 proposals. As visualized in Fig.~\ref{fig:interface_mismatch}, this loss is not limited to complete proposal failure: crack evidence can also be suppressed or spatially reshaped during candidate-mask formation. This diagnostic is used only to identify the representation interface and does not use target labels during SERD-DQ inference. Full thresholds and counting rules are given in~\ref{app:mismatch_protocol}.

\subsection{Semantic-Edge Response Decoding}
\label{sec:serd_decode}

The semantic response provides crack-specific evidence but may remain spatially diffuse near weak boundaries. Image gradients provide complementary boundary information but lack semantics. SERD combines the two without introducing trainable parameters. Let $I_g$ be the normalized grayscale image. A fixed Sobel operator gives
\begin{equation}
E=\operatorname{Norm}\!\left(\sqrt{G_x^2+G_y^2}\right).
\end{equation}
The semantic-edge response is then decoded by
\begin{equation}
R_e=\operatorname{Norm}\!\left(R_q\odot(1+E)\right).
\end{equation}
The multiplicative form preserves semantic gating: a strong image edge cannot independently create a crack response when $R_q$ is negligible. Instead, edge evidence locally reshapes the response where crack semantics already exist, providing sharper transverse cues for response-domain boundary localization. Alternative structural priors and their settings are described in~\ref{app:decoder_details}.

\subsection{Response-Domain Crack Geometry Estimation}
\label{sec:geometry_estimation}

The decoded response $R_e$ is converted into an explicit geometric description without generating a crack mask. SERD-DQ first traces elongated response ridges and links them into ordered crack branches $\Gamma=\{\mathbf p_i\}_{i=1}^{N}$. At each center point, a local tangent direction and its unit normal $\mathbf n_i$ are estimated. A transverse profile is then sampled along
\begin{equation}
\mathbf x_i(d)=\mathbf p_i+d\mathbf n_i.
\end{equation}
The centerline provides the longitudinal crack trajectory, whereas the transverse profile contains the evidence required to localize the two crack sides. Left and right boundary positions are determined jointly from semantic support, response attenuation, and local structural gradients. If the distances from the center point to the two boundaries are $l_i$ and $r_i$, the local crack width is
\begin{equation}
w_i=l_i+r_i.
\end{equation}
Neighboring cross-sections are constrained along each branch to suppress isolated boundary jumps while retaining gradual width variation. The exact ridge detector, branch-linking rule, candidate generation, boundary score, and recentering procedure are implementation details and are given in~\ref{app:quant_protocol}.

\begin{algorithm}[t]
\caption{SERD-DQ for Direct Crack Quantification}
\label{alg:serd_dq}
\KwIn{Image $I$, prompt $q=\texttt{crack}$, frozen SAM3 $\Phi$}
\KwOut{$L$, $\overline{W}$, $W_{\max}$, $A$}

Extract semantic response:
$R_q=\Phi_{\mathrm{resp}}(I,q)$\;

Compute edge field $E$ and decode
$R_e=\mathcal{D}(R_q,E)$\;

Trace response ridges to obtain crack centerline $\Gamma$\;

Estimate local width profile $w(s)$ along the normal direction of $\Gamma$
using $R_e$ and $E$\;

Compute
\[
L=\int_{\Gamma}ds,\qquad
\overline{W}=\frac{1}{L}\int_{\Gamma}w(s)\,ds,
\]
\[
W_{\max}=\max_{s\in\Gamma}w(s),\qquad
A=\int_{\Gamma}w(s)\,ds;
\]

\KwRet{$L,\overline{W},W_{\max},A$}\;
\end{algorithm}

\subsection{SERD-DQ: Direct Crack Quantification}
\label{sec:serd_dq}

SERD-DQ represents each crack by the response-domain centerline $\Gamma$ and its width profile $w(s)$, and derives the requested geometric quantities directly from this representation. The crack length is computed as
\begin{equation}
L=\sum_{i=1}^{N-1}\|\mathbf p_{i+1}-\mathbf p_i\|_2.
\end{equation}
The mean and maximum widths are
\begin{equation}
\overline W=\frac{1}{N}\sum_{i=1}^{N}w_i,
\qquad
W_{\max}=\max_i w_i,
\end{equation}
and the crack area is approximated by integrating the width profile along the traced centerline,
\begin{equation}
A\approx\sum_{i=1}^{N-1}\frac{w_i+w_{i+1}}{2}\|\mathbf p_{i+1}-\mathbf p_i\|_2.
\end{equation}
These quantities are reported in pixels and pixels squared in the present benchmark; physical units can be obtained when a valid image-scale calibration is available. Crucially, no predicted mask, predicted-mask skeleton, or predicted-mask distance transform enters the SERD-DQ path. SERD-DQ therefore preserves continuous semantic and structural evidence until the final local geometric decision. Algorithm~\ref{alg:serd_dq} summarizes the complete SERD-DQ workflow, which directly converts the decoded semantic-edge response into crack length, width, and area measurements without relying on an intermediate predicted mask.

\begin{figure}[t]
    \centering
    \includegraphics[width=0.9\linewidth]{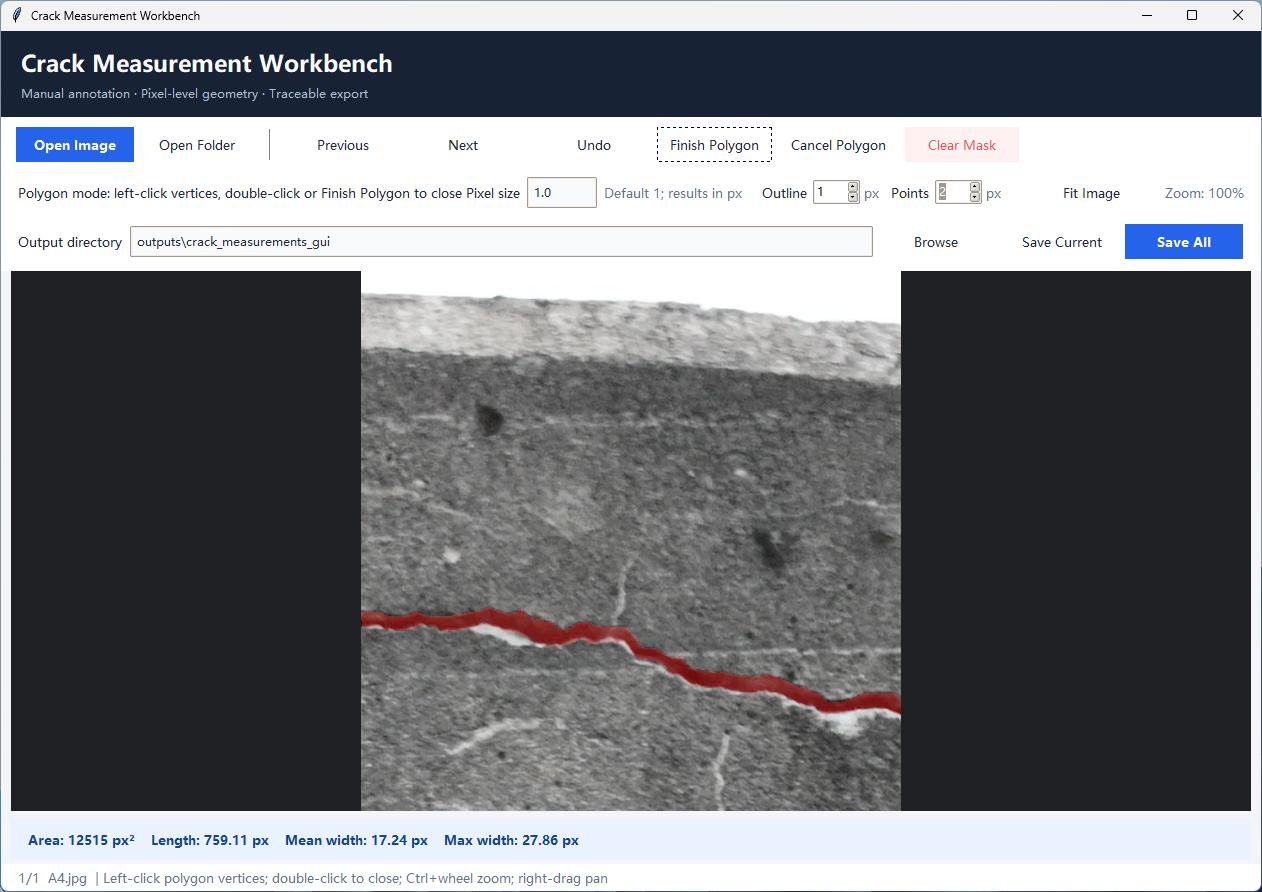}
    \caption{\textbf{Interface of the developed Crack Measurement Workbench for generating manual crack-quantification references.} Domain experts manually delineate the crack region, and the software automatically derives the corresponding pixel-domain geometric quantities, including crack length, mean width, maximum width, and area. These values are used as reference measurements in the crack-quantification experiments.}
\label{fig:measurement_workbench}
\end{figure}

\section{Experimental Setup}
\label{sec:setup}

\subsection{Datasets and Evaluation Scope}

Crack500~\citep{yang2020fphbn}, CamCrack789~\citep{zhu2024camcrack789}, CrackMap~\citep{katsamenis2023crackmap}, DeepCrack~\citep{liu2019deepcrack}, TUT~\citep{liu2024crackscf}, and OmniCrack30k~\citep{benz2024omnicrack30k} are used for segmentation-based validation of crack-response modeling. The datasets cover pavement, concrete, brick, tile, metal, pipe, and multi-source imagery with substantial changes in material, scale, illumination, texture, and morphology. Official train/test partitions are retained when available. Target-domain annotations are never used to update SAM3 or any learned component.

For crack quantification, we construct an image-domain measurement benchmark by randomly sampling crack-bearing images from the test sets of five public datasets rather than collecting a new image set. Sampling follows a fixed pseudorandom draw at a nominal ratio of 5\%, with a minimum of 10 images retained for each domain to preserve cross-domain diversity while keeping the manual measurement workload controllable. Accordingly, $[34, 10, 10, 10, 14]$ images were selected from Crack500, CrackMap, CamCrack789, DeepCrack, and TUT, respectively, resulting in 78 measured images out of 1,244 test images. Each selected image was manually delineated by a domain expert using the developed Crack Measurement Workbench, after which the software automatically computed the pixel-domain crack length, mean width, maximum width, and area. These measurements were used as the manual reference values for evaluating the crack-quantification methods.

\subsection{Pixel-Level Crack Measurement Software and Manual References}
\label{sec:measurement_tool}

To obtain reference measurements that are independent of model predictions, we develop a pixel-level crack measurement tool for manual annotation and automatic geometry calculation. For each sampled image, a domain expert delineates the target crack using the interactive annotation interface, as shown in Fig.~\ref{fig:measurement_workbench}. The software then automatically computes crack length, mean width, maximum width, and area in the image coordinate system, and stores these quantities as the manual reference geometry for quantitative evaluation. Specifically, for an image with a known spatial calibration factor $s_i$ (e.g., mm/pixel), pixel-domain measurements are converted to physical quantities as
\begin{equation}
L_i^{\mathrm{phy}}=s_i L_i^{\mathrm{pix}}, \qquad
W_i^{\mathrm{phy}}=s_i W_i^{\mathrm{pix}}, \qquad
A_i^{\mathrm{phy}}=s_i^2 A_i^{\mathrm{pix}}.
\end{equation}
Thus, the same annotation and measurement procedure can directly produce physical crack dimensions when camera calibration or a scene scale reference is available; otherwise, measurements are reported in pixels and pixels squared.

The tool is used only to construct evaluation references and does not participate in SERD inference. This separation ensures that all compared methods are evaluated against the same manually established crack geometry and allows us to examine whether an intermediate predicted segmentation mask is necessary for accurate crack measurement. Detailed annotation conventions, branch handling, geometric calculations, and scale conversion are provided in~\ref{app:measurement_reference}.

\subsection{Comparison Methods}

\begin{figure}
    \centering
    \includegraphics[width=1\linewidth]{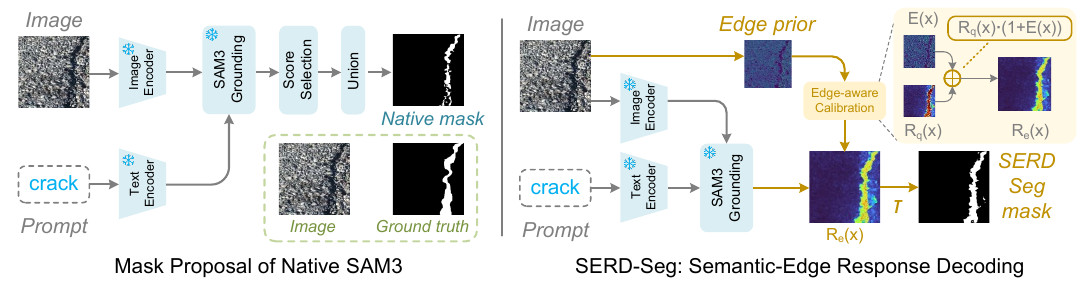}
    \caption{\textbf{Comparison between the native SAM3 mask-generation pathway and the SERD-Seg response readout.} Native SAM3 generates the final crack mask through grounding, score-based proposal selection, and mask union, whereas SERD first calibrates the prompt-conditioned semantic response $R_q$ with the Sobel edge prior $E$ to obtain the decoded response $R_e$. For response-modeling validation, $R_e$ is thresholded by $\tau$ to form the SERD-Seg mask. This readout is used only in the segmentation experiments to assess how effectively the continuous response preserves crack-region evidence.}
\label{fig:serd_seg}
\end{figure}

For response-modeling validation, supervised segmentation references include RINDNet~\citep{pu2021rindnet}, DTrCNet~\citep{xiang2023dtrcnet}, SCSegamba~\citep{liu2025scsegamba}, and MixerCSeg~\citep{zhao2026mixercseg}. Training-free/open-vocabulary references include LaVG~\citep{kang2024lavg}, NACLIP~\citep{hajimiri2025naclip}, CorrCLIP~\citep{zhang2025corrclip}, ReME~\citep{xuan2025reme}, RF-CLIP~\citep{li2026rfclip}, and native SAM3~\citep{carion2025sam3}. The raw SAM3 semantic response is also included to distinguish the effect of response exposure from semantic-edge decoding. To expose the crack-modeling capability of the decoded continuous response through standard segmentation benchmarks, we additionally define an evaluation-only binary readout
\begin{equation}
M_{\mathrm{Seg}}(x)=\mathbbm{1}[R_e(x)\geq\tau],
\end{equation}
which is denoted \textbf{SERD-Seg}. SERD-Seg contains no learned decoder and is not part of the SERD-DQ measurement path; it is introduced only as a benchmark-compatible interface for observing whether $R_e$ preserves crack regions, boundaries, and centerlines. Figure~\ref{fig:serd_seg} illustrates the difference between the native SAM3 mask-generation pathway and the SERD-Seg evaluation readout, where the decoded semantic-edge response $R_e$ is directly thresholded to expose the crack structures retained in the continuous response.

For crack quantification, the controlled comparison focuses on the measurement interface. Native SAM3 and SERD-Seg first produce binary masks and then use the same deterministic mask-to-geometry extractor. SERD-DQ instead estimates crack geometry directly from the continuous decoded response. Using the same reference measurements and the same downstream extractor for all mask-based baselines isolates whether bypassing the predicted mask improves geometric accuracy.

\subsection{Evaluation Metrics}

For response-modeling validation, we use foreground Crack IoU, mIoU, F1, Precision, Recall, and predicted-area ratio (PAR) after a temporary binary readout is formed from the continuous response. Boundary F1 and Boundary IoU measure contour alignment, while clDice~\citep{shit2021cldice} evaluates centerline preservation. Together, these metrics probe three properties required by crack measurement---regional support, transverse boundary localization, and longitudinal continuity---rather than treating segmentation accuracy as the final objective. Detailed metric conventions are provided in~\ref{app:metric_details}.

For quantification, we evaluate four complementary aspects: local width accuracy, longitudinal length accuracy, overall area accuracy, and the recovery of crack geometry discarded by the native SAM3 mask. For the $i$-th image, let $\mathcal S_i^{gt}$ and $\hat{\mathcal S}_i$ denote the GT and predicted crack centerlines, respectively.

\paragraph{\textbf{Width mean absolute error (WMAE)}}
For each GT centerline point $\mathbf p_{ij}^{gt}$, we find its nearest point on $\hat{\mathcal S}_i$, denoted by $\pi_i(\mathbf p_{ij}^{gt})$. Only spatially valid correspondences are retained:
\begin{equation}
\mathcal V_i=
\left\{
j:
d(\mathbf p_{ij}^{gt},\hat{\mathcal S}_i)\leq\delta_w
\right\},
\qquad \delta_w=5~\text{pixels}.
\end{equation}
The width error is then
\begin{equation}
\mathrm{WMAE}_i=
\frac{1}{|\mathcal V_i|}
\sum_{j\in\mathcal V_i}
\left|
\hat w\!\left(\pi_i(\mathbf p_{ij}^{gt})\right)-w_{ij}^{gt}
\right|,
\qquad
\mathrm{WMAE}=
\frac{1}{N_{\mathrm{valid}}}
\sum_{i\in\mathcal I_{\mathrm{valid}}}\mathrm{WMAE}_i .
\end{equation}
WMAE is reported in pixels, where a lower value indicates more accurate local width estimation.

\paragraph{\textbf{Length relative error (LRE)}}
The predicted crack length is obtained by accumulating the arc lengths of all recovered branches:
\begin{equation}
\hat L_i=
\sum_b\sum_j
\left\|
\hat{\mathbf p}_{i,b,j+1}-\hat{\mathbf p}_{i,b,j}
\right\|_2 .
\end{equation}
The image- and dataset-level relative errors are
\begin{equation}
\mathrm{LRE}_i=
\frac{|\hat L_i-L_i^{gt}|}{L_i^{gt}},
\qquad
\mathrm{LRE}=
\frac{1}{N}\sum_{i=1}^{N}\mathrm{LRE}_i .
\end{equation}
LRE is reported as a percentage, with a lower value indicating better longitudinal geometry estimation.

\paragraph{\textbf{Area relative error (ARE)}}
The crack area is directly estimated from the recovered centerline and local width profile:
\begin{equation}
\hat A_i=
\sum_b\sum_j
\frac{\hat w_{i,b,j}+\hat w_{i,b,j+1}}{2}
\left\|
\hat{\mathbf p}_{i,b,j+1}-\hat{\mathbf p}_{i,b,j}
\right\|_2 .
\end{equation}
Accordingly,
\begin{equation}
\mathrm{ARE}_i=
\frac{|\hat A_i-A_i^{gt}|}{A_i^{gt}},
\qquad
\mathrm{ARE}=
\frac{1}{N}\sum_{i=1}^{N}\mathrm{ARE}_i .
\end{equation}
ARE is reported as a percentage, where a lower value indicates more accurate overall crack geometry estimation.

\paragraph{\textbf{Latent geometry recovery rate (LGR)}}
To specifically examine whether SERD-DQ can recover crack evidence discarded by the native SAM3 mask, we define the latent GT geometry as
\begin{equation}
\mathcal S_i^{\mathrm{latent}}
=
\left\{
\mathbf p\in\mathcal S_i^{gt}:
d(\mathbf p,M_i^{\mathrm{native}})>\delta_m,\;
R_{q,i}(\mathbf p)\geq\tau_r
\right\},
\end{equation}
where $\delta_m=2$ pixels and $\tau_r=0.4$. The recovery rate is
\begin{equation}
\mathrm{LGR}_i=
\frac{
\sum_{\mathbf p\in\mathcal S_i^{\mathrm{latent}}}
\mathbbm{1}
\!\left[
d(\mathbf p,\hat{\mathcal S}_i)\leq\delta_l
\right]
}{
|\mathcal S_i^{\mathrm{latent}}|
},
\qquad \delta_l=2~\text{pixels},
\end{equation}
and the dataset-level result is
\begin{equation}
\mathrm{LGR}=
\frac{1}{N_{\mathrm{latent}}}
\sum_{i\in\mathcal I_{\mathrm{latent}}}
\mathrm{LGR}_i .
\end{equation}
A higher LGR indicates that more crack geometry lost at the native SAM3 mask interface is successfully recovered from the internal semantic response.

\subsection{Implementation Details}

All SERD experiments use the official SAM3 image model with frozen parameters and the fixed text prompt \texttt{crack}. Images whose longer side exceeds 1024 pixels are downscaled while preserving aspect ratio before the official SAM3 processor is applied. The semantic response is sigmoid-activated, restored to the evaluation resolution, and normalized independently per image. SERD uses fixed $3\times3$ Sobel kernels and introduces no learned parameters.

For the segmentation-based response validation, the default decoded-response threshold is $\tau=0.45$, selected using the CamCrack789 training split in the default source-to-target protocol and then frozen. This threshold is used only to expose a binary evaluation readout and is not part of SERD-DQ. The non-learned SERD-DQ geometric parameters are fixed once and shared across datasets. Code-level response extraction, geometric tracing, and optional decoder settings are provided in the appendices rather than the main Method section. Experiments are implemented in Python/PyTorch on Ubuntu 22.04 with an NVIDIA RTX A2000 12~GB GPU and an Intel Core i9-11900K CPU.

\section{Experimental Results}
\label{sec:results}

\subsection{Crack Segmentation Comparison for Response-Modeling Validation}
\label{sec:seg_results}

SERD-Seg is used here only as a diagnostic readout of the continuous field $R_e$. If the response represents crack geometry faithfully, a simple monotonic threshold should retain foreground extent, boundary position, and centerline continuity across datasets. We therefore evaluate the readout with established segmentation metrics, while reserving the direct measurement metrics for Section~\ref{sec:quant_comparison}.

\begin{table}[t]
\centering
\small
\setlength{\tabcolsep}{2pt}
\caption{Source-to-target Crack IoU (\%) with CamCrack789 as the source domain. The operating threshold of each method is selected using the CamCrack789 training split and then frozen. The remaining five test sets are target domains, and Avg. denotes their unweighted arithmetic mean. The CamCrack789 column reports source-domain test performance and is excluded from Avg. Supervised methods are task-specific references and are not directly comparable under the frozen-model protocol. Each supervised method is trained on the official training split of the corresponding dataset and evaluated on its test split, following the training configuration of MixerCSeg. $\Delta$ denotes the absolute improvement of SERD-Seg over native SAM3 on each target dataset and on their unweighted average; the source-domain difference is omitted.}
\label{tab1}
\resizebox{\textwidth}{!}{%
\begin{tabular}{@{}l|c|ccccc|c@{}}
\toprule
\textit{\textbf{Supervised methods}} & \textbf{CamCrack789} & \textbf{Crack500} & \textbf{CrackMap} & \textbf{DeepCrack} & \textbf{TUT} & \textbf{OmniCrack30k} & \textbf{Avg.} \\ \midrule

\multicolumn{1}{l|}{RINDNet (2021)} & \multicolumn{1}{c|}{63.93} & 50.66 & 49.51 & 68.39 & 55.17 & \multicolumn{1}{c|}{29.03} & 50.55 \\
\multicolumn{1}{l|}{DTrCNet (2023)} & \multicolumn{1}{c|}{64.17} & 55.40 & 57.14 & 73.73 & 62.10 & \multicolumn{1}{c|}{32.68} & 56.21 \\
\multicolumn{1}{l|}{SCSegamba (2025)} & \multicolumn{1}{c|}{66.65} & 58.43 & 62.20 & 81.22 & 70.23 & \multicolumn{1}{c|}{36.89} & 61.79 \\
\multicolumn{1}{l|}{MixerCSeg (2026)} & \multicolumn{1}{c|}{69.20} & 59.37 & 63.87 & 83.24 & 64.62 & \multicolumn{1}{c|}{34.06} & 61.03 \\ \midrule

\textit{\textbf{Training-free methods}} & \multicolumn{1}{l|}{\textbf{Source domain}} & \multicolumn{5}{c|}{\textbf{Target domain}} & \textbf{Avg.} \\ \midrule
\multicolumn{1}{l|}{LaVG (ECCV'24)} & \multicolumn{1}{c|}{47.59} & 27.88 & 39.37 & 54.18 & 43.10 & \multicolumn{1}{c|}{18.53} & 36.61 \\
\multicolumn{1}{l|}{NACLIP (WACV'25)} & \multicolumn{1}{c|}{48.33} & 28.31 & 39.98 & 55.02 & 43.76 & \multicolumn{1}{c|}{18.82} & 37.18 \\
\multicolumn{1}{l|}{CorrCLIP (ICCV'25)} & \multicolumn{1}{c|}{65.75} & 38.51 & 54.38 & 74.84 & 59.54 & \multicolumn{1}{c|}{25.60} & 50.57 \\
\multicolumn{1}{l|}{ReME (ICCV'25)} & \multicolumn{1}{c|}{67.71} & 39.66 & 56.01 & 77.08 & 61.32 & \multicolumn{1}{c|}{26.36} & 52.09 \\
\multicolumn{1}{l|}{RF-CLIP (AAAI'26)} & \multicolumn{1}{c|}{55.81} & 32.69 & 46.16 & 63.53 & 50.54 & \multicolumn{1}{c|}{21.73} & 42.93 \\
\multicolumn{1}{l|}{SAM3 (ICLR'26)} & \multicolumn{1}{c|}{70.87} & 44.14 & 58.36 & 80.33 & 64.88 & \multicolumn{1}{c|}{23.95} & 54.33 \\
\multicolumn{1}{l|}{SERD-Seg} & \multicolumn{1}{c|}{\textbf{71.34}} & \textbf{53.08} & \textbf{62.78} & \textbf{82.34} & \textbf{66.07} & \multicolumn{1}{c|}{\textbf{25.73}} & \textbf{58.00} \\
\rowcolor{gray!15} \multicolumn{1}{l|}{$\Delta$} & \multicolumn{1}{c|}{-} & \textcolor{ForestGreen}{+8.94} & \textcolor{ForestGreen}{+4.42} & \textcolor{ForestGreen}{+2.01} & \textcolor{ForestGreen}{+1.19} & \multicolumn{1}{c|}{\textcolor{ForestGreen}{+1.78}} & \textcolor{ForestGreen}{+3.67} \\ \bottomrule
\end{tabular}%
}
\end{table}

\begin{table}[t]
\centering
\caption{Average performance of SAM3 and SERD-Seg under the unified operating-point protocol on the same five target-domain test sets as Table~\ref{tab1}. Each value is the unweighted arithmetic mean of the corresponding dataset-level metric. Crack IoU, mIoU, F1, Recall (R), and Precision (P) are percentages. PAR is the ratio of predicted crack area to ground-truth crack area.}
\label{tab2}
\begin{tabular}{ccccccc}
\toprule
\multicolumn{1}{l}{\textbf{Method}} & \textbf{Crack IoU} & \textbf{mIoU} & \textbf{F1} & \textbf{R} & \textbf{P} & \textbf{PAR} \\ \midrule
SAM3 & 54.33 & 76.28 & 64.90 & 69.74 & 66.19 & 0.944 \\
SERD-Seg & 58.00 & 78.04 & 69.53 & 73.95 & 71.83 & 1.098 \\
\rowcolor{gray!15}$\Delta$ & \textcolor{ForestGreen}{+3.67} & \textcolor{ForestGreen}{+1.76} & \textcolor{ForestGreen}{+4.63} & \textcolor{ForestGreen}{+4.21} & \textcolor{ForestGreen}{+5.64} & - \\ \bottomrule
\end{tabular}
\end{table}

\begin{figure}[t]
    \centering
    \includegraphics[width=1\linewidth]{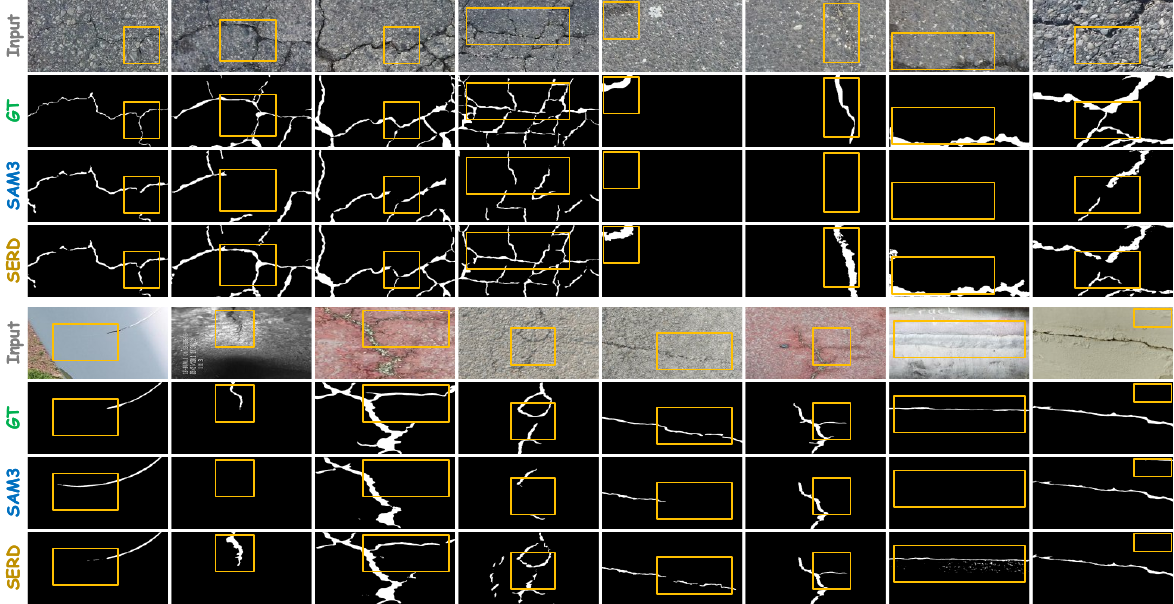}
    \caption{Segmentation-based visualization for validating crack modeling in the decoded response. Each column presents one sample, and the rows show the input image, ground truth, native SAM3 mask, and the SERD-Seg evaluation readout. The marked regions indicate weak branches, disconnected paths, boundary dilation, and crack-like distractors.}
    \label{fig4}
\end{figure}

\begin{table}[t]
\centering
\small
\setlength{\tabcolsep}{4pt}
\caption{Structural response validation using binary readouts from native SAM3, the raw semantic Response, and SERD-Seg. Boundary F1 (B-F1) and Boundary IoU (B-IoU) probe contour alignment, while clDice probes crack-centerline preservation.}
\label{tab3}
\resizebox{\textwidth}{!}{%
\begin{tabular}{@{}lcccccccccccc@{}}
\toprule
\multirow{2}{*}{\textbf{Variant}} & \multicolumn{4}{c}{\textbf{Crack500}} & \multicolumn{4}{c}{\textbf{CrackMap}} & \multicolumn{4}{c}{\textbf{DeepCrack}} \\ \cmidrule(lr){2-5}\cmidrule(lr){6-9}\cmidrule(lr){10-13}
 & \textbf{IoU} & \textbf{B-F1} & \textbf{B-IoU} & \textbf{clDice} & \textbf{IoU} & \textbf{B-F1} & \textbf{B-IoU} & \textbf{clDice} & \textbf{IoU} & \textbf{B-F1} & \textbf{B-IoU} & \textbf{clDice} \\ \midrule
SAM3 & 44.14 & 36.23 & 24.89 & 60.17 & 58.36 & 76.03 & 65.36 & 80.73 & 80.33 & 83.13 & 72.42 & 92.29 \\
Response & \textbf{54.68} & 41.64 & 28.03 & \textbf{72.75} & \textbf{63.89} & 80.26 & 67.91 & \textbf{88.55} & 80.45 & 82.87 & 71.53 & \textbf{94.12} \\
SERD-Seg & 53.08 & \textbf{42.16} & \textbf{28.62} & 70.74 & 62.78 & \textbf{81.56} & \textbf{69.46} & 86.98 & \textbf{82.34} & \textbf{84.94} & \textbf{74.57} & 93.65 \\ \bottomrule
\end{tabular}
}
\end{table}

\paragraph{\textbf{Region-level response fidelity}}
Table~\ref{tab1} compares the SERD-Seg readout with supervised and training-free segmentation references under the CamCrack789-to-five-target protocol. The purpose of this comparison is to test whether the decoded response can recover crack regions under a simple threshold, not to position segmentation as the final task. SERD-Seg reaches 58.00\% mean target-domain Crack IoU, 3.67 percentage points above the native SAM3 mask. The largest difference occurs on Crack500, where Crack IoU increases from 44.14\% to 53.08\%. Because the SAM3 weights and prompt are unchanged and the readout contains no learned decoder, these results provide direct evidence that the decoded response itself models crack foreground structure more faithfully than the native final-mask interface.

Table~\ref{tab2} further shows that the response improvement is not confined to a single overlap statistic. After the same binary readout, the decoded response improves Crack IoU, mIoU, F1, Recall, and Precision simultaneously. This consistency supports the interpretation that SERD changes the underlying crack evidence rather than exploiting one threshold-specific metric. At the same time, PAR moves from slight under-coverage to moderate over-coverage, illustrating why this validation stage cannot replace direct measurement evaluation: a binary region may obtain favorable overlap while its local boundaries remain displaced.

\paragraph{\textbf{Boundary and centerline response fidelity}}
For crack measurement, boundary alignment and centerline continuity are more directly relevant than region overlap alone. Table~\ref{tab3} therefore uses segmentation-derived structural metrics as probes of the response geometry. The raw semantic Response strongly improves clDice relative to native SAM3 proposals, whereas semantic-edge decoding generally provides the strongest boundary metrics. This division is consistent with the SERD-DQ design: semantic responses preserve longitudinal crack evidence, while the Sobel field supplies additional transverse localization cues for width estimation.

\begin{figure}[t]
    \centering
    \includegraphics[width=1\linewidth]{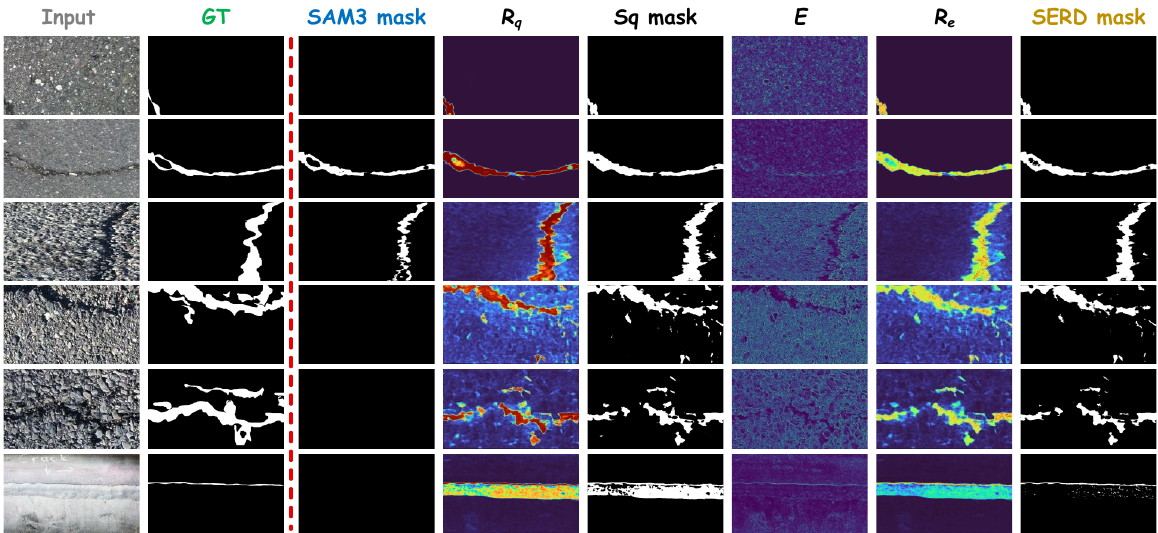}
    \caption{Evaluation-only binarization used to expose the crack-modeling quality of the SERD response field. From left to right, each row shows the input image, ground truth, native SAM3 proposal, normalized semantic response $R_q$, directly thresholded response mask, Sobel edge prior $E$, calibrated response $R_e$, and the SERD-Seg readout. The binary outputs are used only for representation validation and do not enter SERD-DQ.}
    \label{fig5}
\end{figure}

\subsection{Crack Quantification Comparison}
\label{sec:quant_comparison}

Table~\ref{tab:direct_quant} compares the crack measurement performance of native SAM3, SERD-Seg, and SERD-DQ. For the first two methods, geometric quantities are extracted from their binary masks using the same downstream measurement procedure, whereas SERD-DQ directly estimates crack geometry from the decoded semantic-edge response without generating an intermediate mask. This comparison therefore separates the benefit of improved crack representation from that of direct response-domain quantification.

\begin{table}[t]
\centering
\small
\caption{\textbf{Primary image-domain crack-quantification results.} For mask-based methods, geometric quantities are extracted from the predicted masks using the same downstream measurement procedure. WMAE is reported in pixels, LRE and ARE in percentages, and LGR as a ratio in $[0,1]$. LGR is not reported for the native SAM3 mask because latent geometry is defined with respect to crack structures missed by this mask.}
\label{tab:direct_quant}
\resizebox{\textwidth}{!}{%
\begin{tabular}{lccccc}
\toprule
\textbf{Method} & \textbf{Explicit mask} & \textbf{WMAE $\downarrow$} & \textbf{LRE $\downarrow$} & \textbf{ARE $\downarrow$} & \textbf{LGR $\uparrow$} \\ \midrule
SAM3 mask $\rightarrow$ geometry & \checkmark & 6.072 & 22.245 & 33.921 & N/A \\
SERD-Seg $\rightarrow$ geometry & \checkmark & 5.822 & 18.455 & 28.516 & 0.363 \\
\textbf{SERD-DQ (ours)} & $\times$ & \textbf{5.547} & \textbf{17.355} & \textbf{26.228} & \textbf{0.394} \\ \bottomrule
\end{tabular}%
}
\end{table}

As shown in Table~\ref{tab:direct_quant}, replacing the native SAM3 mask with the SERD-Seg mask consistently improves all measurement metrics. WMAE decreases from 6.072 to 5.822 pixels, while LRE and ARE decrease from 22.245\% and 33.921\% to 18.455\% and 28.516\%, respectively. These results are consistent with the segmentation experiments in Section~\ref{sec:seg_results} and further indicate that the decoded semantic-edge response provides a more reliable geometric representation of cracks than the native SAM3 output mask.

\begin{figure}[t]
    \centering
    \includegraphics[width=1\linewidth]{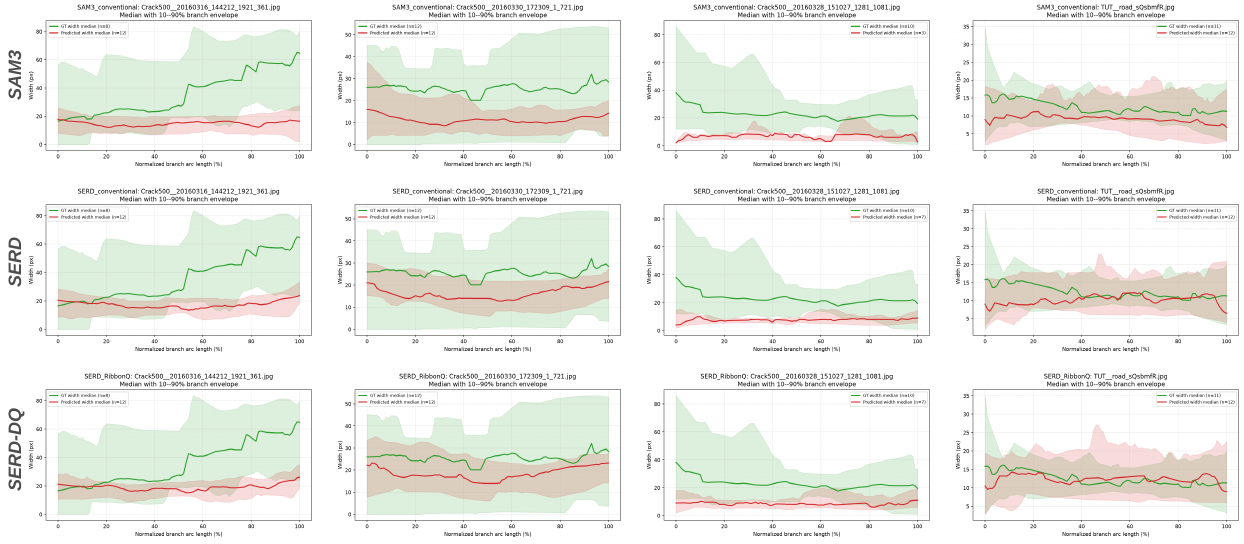}
    \caption{\textbf{Comparison of local crack-width profiles.} Rows correspond to SAM3 mask-based quantification, SERD-Seg mask-based quantification, and SERD-DQ, while columns show representative crack images from different domains. Green and red curves denote the median reference and predicted width profiles, respectively, and the shaded regions indicate the 10--90\% variation range across valid crack branches. The horizontal axis is the normalized branch arc length and the vertical axis is crack width in pixels. SERD-DQ generally better preserves the spatial variation of crack width by directly measuring geometry from the continuous semantic-edge response.}
    \label{fig:width_profile}
\end{figure}

SERD-DQ further improves the measurement accuracy without relying on explicit mask generation. Compared with native SAM3, SERD-DQ reduces WMAE by 8.65\%, LRE by 21.98\%, and ARE by 22.68\%, achieving 5.547 pixels, 17.355\%, and 26.228\%, respectively. It also consistently outperforms SERD-Seg, reducing WMAE, LRE, and ARE by 4.72\%, 5.96\%, and 8.02\%. More importantly, SERD-DQ achieves an LGR of 0.394, compared with 0.363 for SERD-Seg, showing that direct response-domain geometry estimation recovers more crack structures that are discarded at the native SAM3 mask interface. These results suggest that binarization is not only unnecessary for crack measurement, but can also discard geometric evidence useful for subsequent quantification.

\begin{figure}[t]
    \centering
    \includegraphics[width=0.9\linewidth]{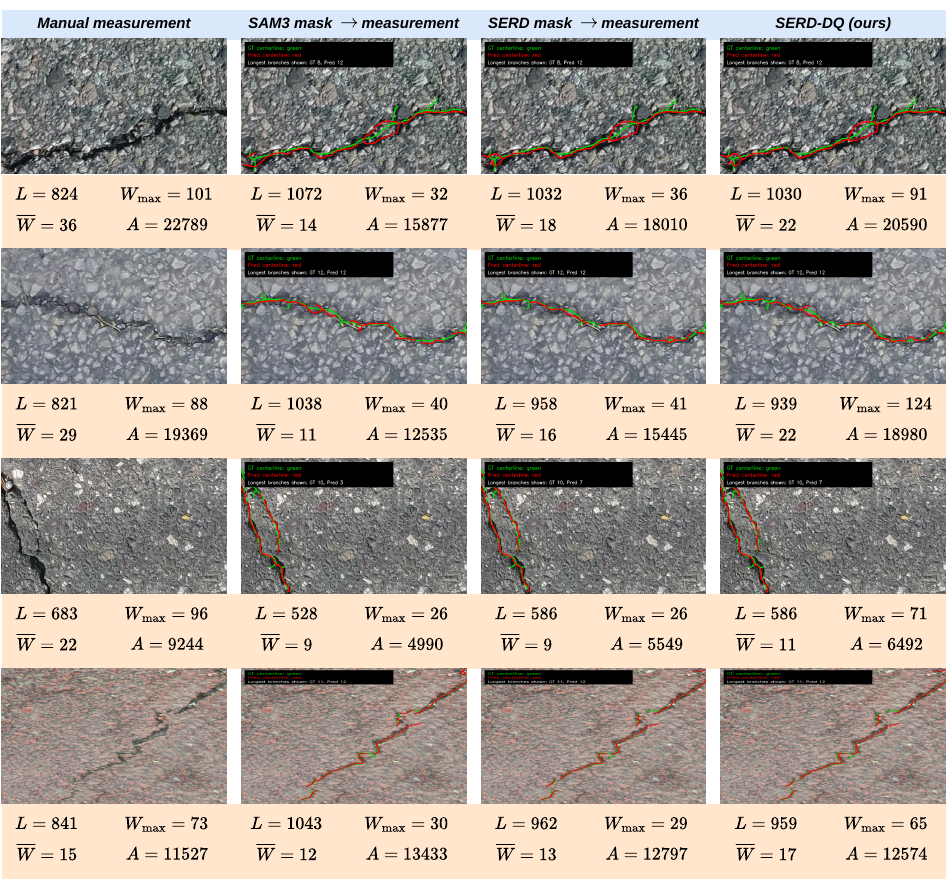}
    \caption{\textbf{Qualitative comparison of crack quantification.} Manually annotated reference measurements, SAM3 mask-based quantification, SERD-Seg mask-based quantification, and the proposed SERD-DQ are compared on representative crack images. Green and red curves denote the reference and predicted crack centerlines, respectively. For each example, $L$, $W_{\max}$, $\overline{W}$, and $A$ denote crack length, maximum width, mean width, and area, respectively. All geometric quantities are measured in the image domain.}
    \label{fig:quant_vis}
\end{figure}

Figure~\ref{fig:quant_vis} shows representative cases in which native SAM3 misses thin branches or contracts local boundaries, leading to centerline and width errors. SERD-Seg recovers part of this geometry, while SERD-DQ retains the continuous response through the final cross-sectional estimate and generally follows the manual reference more closely. Local maximum-width overestimation remains possible in ambiguous regions.

The width profiles in Figure~\ref{fig:width_profile} provide the corresponding local view. Mask-based estimates are sensitive to boundary contraction and fragmentation, whereas SERD-DQ more consistently tracks the spatial variation of the reference width profile. This agrees with the WMAE reduction and directly supports the use of continuous response evidence for transverse geometry.

\subsection{Ablation Study}
\label{sec:ablation}

\paragraph{\textbf{Structural edge prior}}
Table~\ref{tab5} compares Sobel with Canny, Laplacian of Gaussian, and morphological black-hat using the same semantic-response decoding pipeline. Although black-hat gives a marginally higher average IoU, Sobel provides consistently strong regional performance without additional detection thresholds or scale-dependent structuring elements. It is therefore selected as the default deterministic structural prior.

\begin{table}[t]
\centering
\caption{Effect of different structural priors on the SERD-Seg readout. Sobel, Canny, Laplacian of Gaussian (LoG), and morphological black-hat are compared using the same semantic-response extraction and decoding pipeline. Results are averaged over Crack500, CrackMap, and DeepCrack.}
\label{tab5}
\begin{tabular}{lccccc}
\toprule
\multicolumn{1}{l}{\textbf{Edge Type}} & \textbf{IoU} & \textbf{F1} & \textbf{R} & \textbf{P} & \textbf{PAR} \\ \midrule
Sobel & \underline{66.07} & \underline{77.89} & \underline{78.40} & 80.61 & 1.136 \\
Canny & 64.73 & 76.76 & 75.56 & \textbf{81.45} & 1.025 \\
LoG & 65.31 & 77.39 & 77.08 & 81.01 & 1.131 \\
black-hat & \textbf{66.49} & \textbf{78.18} & \textbf{78.45} & \underline{81.05} & 1.140 \\ \bottomrule
\end{tabular}
\end{table}

\paragraph{\textbf{Additional binary decoding operations}}
We additionally examine a weak-response branch and topology filtering. As shown in Table~\ref{tab6}, these operations provide negligible gains over the basic semantic-edge response readout. They are therefore excluded from the main SERD formulation, supporting the use of a simple response decoder rather than a progressively more complex post-processing pipeline.

\begin{table}[t]
\centering
\small
\setlength{\tabcolsep}{3.5pt}
\caption{Ablation study of the decoding design on Crack500, CrackMap, and DeepCrack. A weak semantic-response branch ($W$) and topology filtering ($T$) are added separately to the default SERD-Seg readout. Semantic-edge calibration is retained in all variants so that the effects of the additional components can be evaluated independently. $W$ introduces a low-response threshold and an edge-support gate, whereas $T$ introduces an adaptive high-confidence seed and an eight-connected minimum-area component filter. All associated hyperparameters are fixed as specified in~\ref{app:decoder_details}.}
\label{tab6}
\resizebox{\textwidth}{!}{%
\begin{tabular}{ccccccccccccccccc}
\toprule
\multirow{2}{*}{\textbf{$W$}} & \multirow{2}{*}{\textbf{$T$}} & \multicolumn{5}{c}{\textbf{Crack500}} & \multicolumn{5}{c}{\textbf{CrackMap}} & \multicolumn{5}{c}{\textbf{DeepCrack}} \\ \cmidrule(lr){3-7}\cmidrule(lr){8-12}\cmidrule(lr){13-17}
 &  & \textbf{IoU} & \textbf{F1} & \textbf{R} & \textbf{P} & \textbf{PAR} & \textbf{IoU} & \textbf{F1} & \textbf{R} & \textbf{P} & \textbf{PAR} & \textbf{IoU} & \textbf{F1} & \textbf{R} & \textbf{P} & \textbf{PAR} \\ \midrule
 &  & 53.1 & 66.7 & 65.2 & 74.9 & 1.331 & 62.8 & 76.9 & 77.1 & 78.6 & 1.008 & 82.3 & 90.1 & 92.9 & 88.3 & 1.068 \\
\checkmark &  & 53.1 & 66.7 & 65.3 & 74.7 & 1.339 & 62.8 & 76.9 & 77.3 & 78.5 & 1.013 & 82.4 & 90.1 & 93.2 & 88.1 & 1.073 \\
 & \checkmark & 53.0 & 66.7 & 64.9 & 75.1 & 1.297 & 62.7 & 76.8 & 76.9 & 78.8 & 1.003 & 82.5 & 90.2 & 92.9 & 88.4 & 1.066 \\
\checkmark & \checkmark & 53.1 & 66.7 & 65.0 & 75.0 & 1.300 & 62.8 & 76.9 & 77.1 & 78.7 & 1.007 & 82.5 & 90.2 & 93.2 & 88.2 & 1.071 \\ \bottomrule
\end{tabular}
}
\end{table}

\begin{figure}[t]
    \centering
    \includegraphics[width=0.52\linewidth]{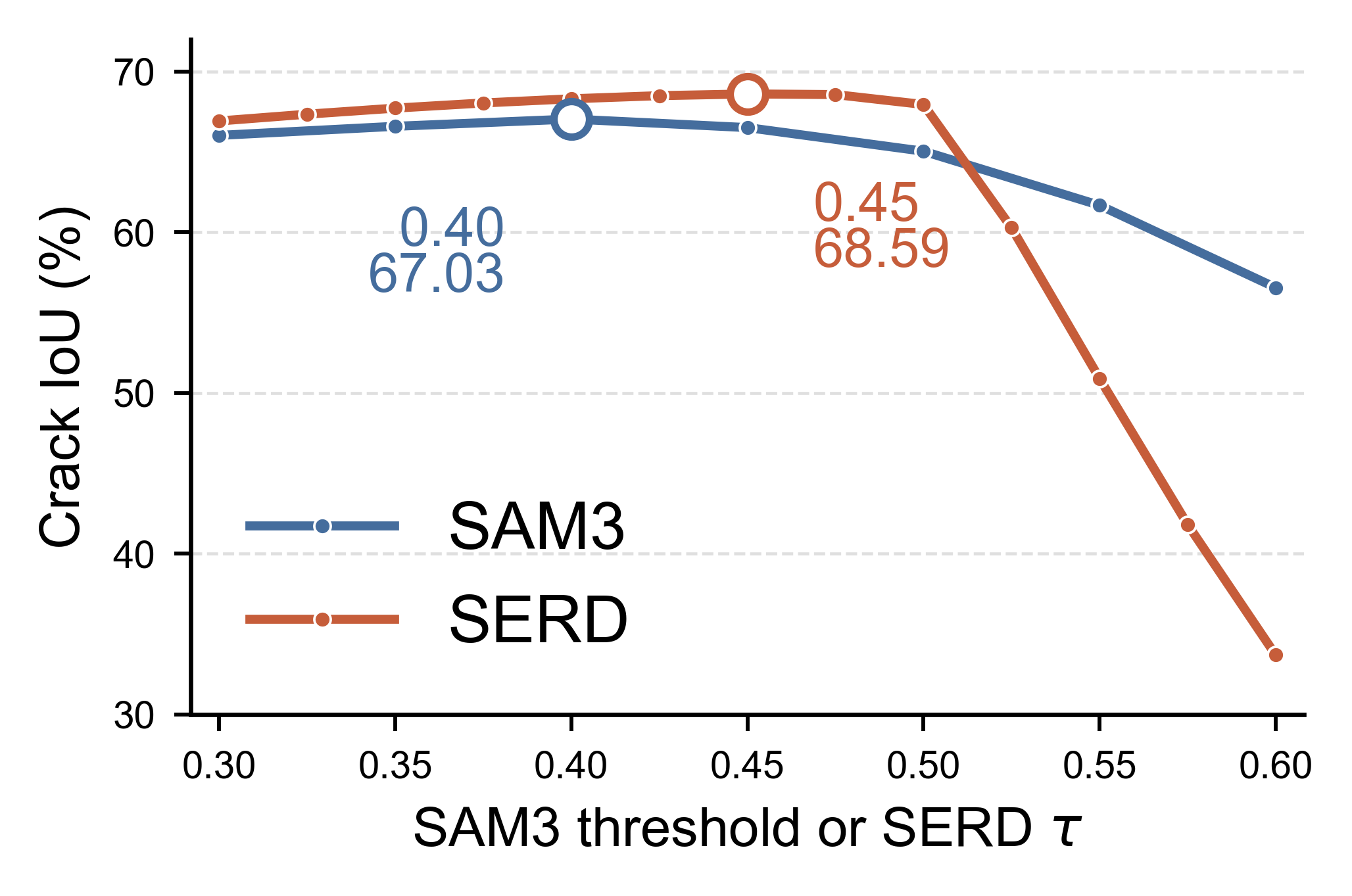}
    \caption{Sensitivity of native SAM3 and SERD-Seg to their respective operating thresholds on the CamCrack789 training split. The enlarged markers indicate the operating thresholds selected on this source split.}
    \label{fig7}
\end{figure}

\paragraph{\textbf{Segmentation operating point}}
Figure~\ref{fig7} reports the threshold sensitivity of native SAM3 and the SERD-Seg evaluation readout on the source split. This operating point is used only for segmentation-based response validation; SERD-DQ does not use the SERD-Seg binarization threshold.

\subsection{Cross-Domain Robustness and Deployment Cost}

Rapid deployment requires the response representation to remain useful when inspection material and acquisition conditions change. Rotating all six public datasets through the source role shows that the SERD-Seg validation readout consistently improves target-domain Precision and Boundary F1 over native SAM3, while the raw Response and SERD-Seg have similar average IoU under some source choices. Interpreted as a representation test, this result indicates that semantic-edge decoding is particularly useful for stabilizing crack-boundary evidence rather than merely increasing foreground coverage.

\begin{table}[t]
\centering
\caption{Cross-domain robustness under different source-domain calibrations. Each dataset is used in turn as the source domain. The reported threshold is selected using only the corresponding source-domain training split, and Crack IoU (IoU), Precision (P), PAR, and Boundary F1 (B-F1) are averaged over the test sets of the other five datasets. Target-domain labels are not used for threshold selection. For SAM3, Thd is the proposal-score threshold; for Response and SERD-Seg, it is the dense-response threshold.}
\label{tab:my-table}
\resizebox{\textwidth}{!}{%
\begin{tabular}{@{}lccccccccccccccc@{}}
\toprule
\textbf{Method} & \multicolumn{5}{c}{\textbf{CamCrack789}} & \multicolumn{5}{c}{\textbf{Crack500}} & \multicolumn{5}{c}{\textbf{CrackMap}} \\ \cmidrule(lr){2-6} \cmidrule(lr){7-11} \cmidrule(lr){12-16} 
 & \textbf{Thd} & \textbf{IoU} & \textbf{P} & \textbf{PAR} & \textbf{B-F1} & \textbf{Thd} & \textbf{IoU} & \textbf{P} & \textbf{PAR} & \textbf{B-F1} & \textbf{Thd} & \textbf{IoU} & \textbf{P} & \textbf{PAR} & \textbf{B-F1} \\ \midrule
SAM3 & 0.4 & 54.33 & 66.19 & 0.944 & 61.23 & 0.3 & 59.43 & 68.64 & 1.404 & 70.35 & 0.3 & 56.42 & 66.16 & 1.342 & 61.69 \\
Response & 0.4 & \textbf{58.40} & 67.83 & 1.324 & 63.72 & 0.3 & 60.91 & 66.92 & 1.920 & 70.15 & 0.3 & 59.12 & 66.55 & 1.902 & 62.49 \\
SERD-Seg & 0.45 & 58.00 & \textbf{71.83} & 1.098 & \textbf{65.12} & 0.3 & \textbf{61.76} & \textbf{69.77} & 1.709 & \textbf{71.93} & 0.3 & \textbf{59.88} & \textbf{69.23} & 1.690 & \textbf{64.16} \\ \midrule
\multirow{2}{*}{\textbf{Method}} & \multicolumn{5}{c}{\textbf{DeepCrack}} & \multicolumn{5}{c}{\textbf{TUT}} & \multicolumn{5}{c}{\textbf{OmniCrack30k}} \\ \cmidrule(lr){2-6} \cmidrule(lr){7-11} \cmidrule(lr){12-16} 
 & \textbf{Thd} & \textbf{IoU} & \textbf{P} & \textbf{PAR} & \textbf{B-F1} & \textbf{Thd} & \textbf{IoU} & \textbf{P} & \textbf{PAR} & \textbf{B-F1} & \textbf{Thd} & \textbf{IoU} & \textbf{P} & \textbf{PAR} & \textbf{B-F1} \\ \midrule
SAM3 & 0.45 & 52.20 & 65.06 & 1.037 & 60.58 & 0.35 & 55.92 & 67.00 & 1.191 & 64.22 & 0.45 & 63.42 & 75.92 & 0.892 & 68.78 \\
Response & 0.45 & \textbf{56.36} & 67.03 & 1.686 & 63.41 & 0.35 & 58.77 & 66.95 & 1.834 & 65.17 & 0.45 & 67.34 & 77.30 & 1.085 & 71.20 \\
SERD-Seg & 0.5 & 55.00 & \textbf{71.86} & 1.248 & \textbf{64.21} & 0.3 & \textbf{59.22} & \textbf{68.64} & 1.701 & \textbf{66.20} & 0.3 & \textbf{67.50} & \textbf{77.57} & 1.081 & \textbf{71.43} \\ \bottomrule
\end{tabular}%
}
\end{table}

\paragraph{\textbf{Deployment cost}}
SERD uses the same single SAM3 forward pass as the native proposal pipeline and introduces no trainable network. The timing experiment in Table~\ref{tab7} shows that response extraction and decoding have no meaningful additional model-inference cost relative to native SAM3 proposal generation. SERD-DQ adds only deterministic response-domain geometry operations after the same forward pass. This computational profile is consistent with the intended rapid-deployment setting, where the primary saving is the removal of target-domain annotation and model optimization rather than a claim of replacing SAM3 with a lightweight backbone.

\begin{table}[t]
    \centering
    \small
    \setlength{\tabcolsep}{5pt}
    \caption{Inference latency of SAM3 and SERD-Seg. Results are averaged over 120 images from six datasets after GPU warm-up. Both methods use identical SAM3 weights, prompts, input resolutions, and CUDA settings. $\Delta$ denotes the latency difference between SERD and SAM3, with negative values indicating lower latency for SERD.}
    \resizebox{\textwidth}{!}{%
    \begin{tabular}{c|cccccc|c}
    \toprule
    \textbf{Method} & \textbf{Crack500} & \textbf{CamCrack789} & \textbf{CrackMap} & \textbf{DeepCrack} & \textbf{TUT} & \textbf{OmniCrack30k} & \textbf{Avg.} \\ \midrule
    SAM3 & 550.1 & 571.1 & 534.7 & 563.4 & 616.9 & 596.5 & 572.1 \\
    SERD-Seg & 534.9 & 545.3 & 538.1 & 546.3 & 574.0 & 565.2 & 550.6 \\
    \rowcolor{gray!15} $\Delta$ & \textcolor{ForestGreen}{-15.2} & \textcolor{ForestGreen}{-25.8} & \textcolor{red}{+3.4} & \textcolor{ForestGreen}{-17.1} & \textcolor{ForestGreen}{-42.9} & \textcolor{ForestGreen}{-31.3} & \textcolor{ForestGreen}{-21.5} \\ \bottomrule
    \end{tabular}
}
    \label{tab7}
\end{table}

\section{Discussion and Limitations}
\label{sec:discussion}

\paragraph{\textbf{Why direct quantification is useful}}
The comparison between the SERD-Seg evaluation readout and SERD-DQ isolates the effect of the intermediate representation while keeping the same decoded response $R_e$. A binary mask is useful for making crack-modeling quality observable through standard benchmarks, but crack measurement ultimately requires a longitudinal trajectory and reliable transverse boundaries. For a thin crack, moving both predicted boundaries by only a few pixels can cause a large relative width error even when region overlap remains acceptable. SERD-DQ therefore preserves continuous response magnitude until the local geometric decision is made, avoiding a global binarization step in the actual measurement path. This does not imply that segmentation is unnecessary for all inspection tasks; it only means that segmentation need not be treated as a compulsory interface when the final objective is geometric measurement.

\paragraph{\textbf{Implications for Engineering Knowledge Formalization}}
Beyond crack measurement, the results highlight a distinction between a \emph{prediction interface} and an \emph{engineering-information interface}. A prediction interface is optimized to express the output required by a recognition task, such as a class label or binary mask. An engineering-information interface instead needs to preserve the attributes that remain relevant to subsequent measurement, reasoning, or model updating. Previous engineering-informatics studies have addressed this issue at different abstraction levels by transforming heterogeneous data into knowledge graphs, digital-twin states, or model-update variables~\citep{teclaw2024bimkg,sun2024hybridtwin,chang2026tunnelcv}. Our experiments identify an analogous issue inside a vision foundation model: the final task output can remove spatial evidence that remains useful for engineering quantification.

SERD-DQ therefore instantiates a more general representation pathway in which raw observations are first expressed as latent computational evidence, this evidence is subsequently organized into structured engineering geometry, and the resulting geometric representation is finally converted into measurable engineering quantities. In contrast to conventional segment-then-measure pipelines, this process does not require the latent evidence to be discretized into a segmentation mask before engineering interpretation. In the present study, the corresponding representations are 
$I\rightarrow R_e\rightarrow\{\Gamma,w(s)\}\rightarrow\{L,\overline W,W_{\max},A\}$. 

The empirical benefit of this formulation should not be interpreted as evidence that segmentation is generally unnecessary. Instead, it suggests that when the downstream task depends strongly on continuity, local boundaries, or continuously varying dimensions, the choice of intermediate representation should be determined by the engineering information that must be preserved rather than solely by the conventional output format of the visual model.

This insight may be relevant to other geometry-sensitive engineering perception problems in which a pretrained model contains spatial evidence richer than its final discrete prediction. The transferable principle is not the crack-specific Sobel calibration or the particular width estimator used here, but the strategy of interrogating internal computational representations, determining which information is lost at the standard output interface, and formalizing the retained evidence into task-relevant engineering variables. Whether this principle extends to other damage morphologies, materials, sensing modalities, and physical quantities requires dedicated validation and remains an important direction for future engineering-informatics research.

\begin{figure}[t]
    \centering
    \includegraphics[width=1\linewidth]{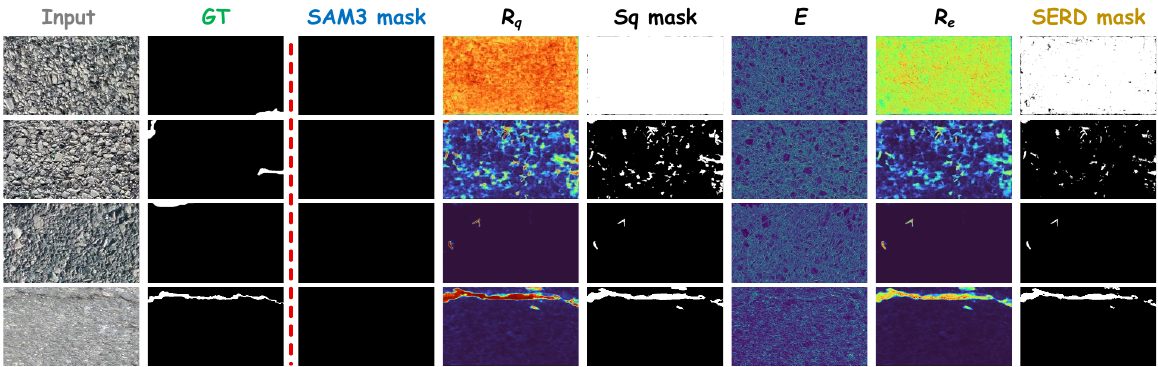}
    \caption{Representative failure cases of the SERD-Seg readout. Weak or diffuse semantic evidence causes missed low-contrast cracks, whereas joints, scratches, shadows, and repetitive textures may receive both semantic and structural support and remain as false positives. SERD-Seg can expose existing evidence but cannot reconstruct absent responses or consistently reject crack-like distractors.}
    \label{fig8}
\end{figure}

\paragraph{\textbf{Limitations}}
SERD can only reorganize evidence already present in the frozen foundation model. If the prompt-conditioned semantic response is absent or extremely weak, the validation readout exposes the omission and SERD-DQ may truncate the centerline. Conversely, joints, scratches, shadows, seams, and repetitive textures can receive both semantic and structural support and create false geometric traces. Direct quantification introduces a further source of error: even when the centerline is correct, a weak or misleading image edge can bias the estimated transverse boundary and therefore the crack width. Figure~\ref{fig8} illustrates representative response failures through the SERD-Seg readout; the same missing or spurious evidence can propagate to direct measurement.

The present quantification benchmark is image based. Length and width are reported in pixels and area in pixels squared because the public datasets do not provide uniform camera calibration. Converting these quantities to millimeters requires an appropriate scale model, and perspective scenes may require spatially varying calibration. In addition, the manually established measurement references inherit operator uncertainty. Future work should therefore evaluate the response-to-measurement formulation on acquisition systems with controlled scale, repeated manual measurements, and physical crack-width references.

\section{Conclusion}

This paper investigates rapidly deployable crack measurement from the internal response representation of a promptable vision foundation model. We show that native SAM3 crack masks exhibit an output-interface mismatch: useful crack evidence remains in the prompt-conditioned semantic response but can be lost or spatially altered during final proposal generation. Based on this observation, SERD decodes the SAM3 response with a fixed Sobel structural field, and SERD-DQ directly estimates crack centerline and transverse geometry to obtain length, mean width, maximum width, and area without generating a predicted binary mask. To verify that the decoded response genuinely models crack geometry, we introduce SERD-Seg only as an experimental binary readout and evaluate it with established region, boundary, and topology metrics; segmentation is therefore used as representation evidence rather than the final objective. A pixel-level manual measurement tool is further used to establish reference geometric quantities on randomly sampled public crack images for direct quantification evaluation. The resulting framework shifts crack analysis from a compulsory segment-then-measure pipeline toward a training-free response-to-measurement alternative that explicitly avoids mask-induced geometric errors.

More broadly, the study suggests that the final discrete output of a visual model should not automatically be regarded as the only interface for formalizing engineering knowledge. When downstream engineering analysis depends on geometric continuity or continuously varying attributes, internal computational responses may provide a more information-preserving intermediate representation. SERD-DQ provides one concrete instance of this principle by transforming latent foundation-model evidence into structured and measurable engineering geometry, while its applicability beyond crack measurement remains to be established in future domain-specific studies.

\section*{Acknowledgment}
This work would not have been possible without financial support from the National Natural Science Foundation of China (Grant Nos. 52178393 and 51578447), the Science and Technology Innovation Team of Shaanxi Innovation Capability Support Plan (2020TD-005), the Xi'an Scientists $\&$ Engineers Workforce Building Project (2024JH-KGDW-0112), and the Key Project of the Natural Science Basic Research Program of Shaanxi Province (2025SYS-SYSZD-049). The authors are also grateful to the editors and anonymous reviewers for their sound and valuable suggestions on the manuscript.

\section*{Declaration of generative AI and AI-assisted technologies in the manuscript preparation process}
During the preparation of this work the authors used ChatGPT (OpenAI) for translation and language polishing. After using this tool/service, the authors reviewed and edited the content as needed and take full responsibility for the content of the manuscript.

\bibliographystyle{elsarticle-num}
\bibliography{SERD_AEI}

\newpage
\appendix

\section{SERD-DQ Implementation Protocol}
\label{app:quant_protocol}

This appendix specifies the deterministic geometry-extraction protocol used by SERD-DQ. Response smoothing, ridge extraction, branch linking, profile sampling, boundary-candidate generation, continuity regularization, recentering, and branch pruning use one fixed non-learned configuration for all evaluated images and datasets; no target-domain labels are used to adjust these settings.

\paragraph{\textbf{Ridge extraction and branch linking}}
Let $H_R(x)$ denote the Hessian of a mildly smoothed $R_e$ and let $\lambda_1(x)\leq\lambda_2(x)$ denote its eigenvalues. For an elongated bright ridge, strong negative curvature is expected across the crack. The ridge support is defined as
\begin{equation}
S_r(x)=R_e(x)\max(0,-\lambda_1(x)).
\end{equation}
The eigenvector associated with $\lambda_1$ defines the local normal $\mathbf n(x)$, and the orthogonal direction defines the local tangent $\mathbf t(x)$. Sub-pixel maxima along $\mathbf n(x)$ are retained as center candidates and linked using spatial proximity, response strength, and tangent consistency. Gap bridging is allowed only when endpoint geometry and intervening response remain compatible.

\paragraph{\textbf{Cross-sectional candidate generation}}
For each resampled center point $\mathbf p_i$, candidate profiles are evaluated over
\begin{equation}
\mathbf x_i(d)=\mathbf p_i+d\mathbf n_i,\qquad d\in[-d_{\max},d_{\max}].
\end{equation}
A candidate section is represented by asymmetric left and right distances $l_i>0$ and $r_i>0$, giving $w_i=l_i+r_i$. The asymmetric form allows recentering when the initial semantic ridge is not exactly centered between the two physical boundaries.

\paragraph{\textbf{Boundary scoring and branch continuity}}
Semantic interval support is measured by the contrast between the mean response inside a candidate section and the response immediately outside it. Response drops at the two endpoints and local edge magnitude provide complementary boundary terms. Multiple candidates are retained at each center point and optimized along a branch with a robust continuity penalty on changes in width and center offset. The selected offsets are smoothed and used to recenter the trajectory before final aggregation. The same candidate ranges and weighting coefficients are used throughout the benchmark, so the direct-quantification path contains no image-wise or dataset-wise parameter selection.

\section{Pixel-Level Manual Measurement Reference Protocol}
\label{app:measurement_reference}

The measurement software is used only to construct reference quantities from the 78 sampled public images described in Section~\ref{sec:setup}. Annotation is performed in the image coordinate system with zoom and pan available for local inspection, while stored polygon coordinates remain tied to the original image pixels. The operator delineates the visually identifiable crack region, including visible branches through intersections; no crack continuation is extrapolated across genuinely occluded or visually indeterminate regions. The finalized annotation is rasterized once and passed to one deterministic reference-geometry routine, which computes centerline length, local width statistics, and crack area under the same convention for every image. The per-dataset sample counts are 34 Crack500, 10 CrackMap, 10 CamCrack789, 10 DeepCrack, and 14 TUT images. These reference annotations are created independently of all model outputs and are never used to tune SERD or SERD-DQ.

When a scalar spatial calibration $s_i$ in mm/pixel is available, the software converts pixel-domain length and width by $s_i$ and area by $s_i^2$. The public datasets used in the benchmark do not provide a uniform physical calibration, so the reported comparison remains in pixels and pixels squared.

\section{Code-Level Extraction of the SAM3 Semantic Response}
\label{app:response_extract}

All response-extraction and SERD-Seg experiments use the official SAM3 image repository at commit \path{5dd401d1c5c1d5c3eedff06d41b77af824517619} and the public SERD implementation at commit \path{0f85b983392c135b63ffb9372b22cb36b9a36379}. The current response-extraction code invokes \path{build_sam3_image_model} with the authorized image checkpoint supplied through \path{--sam_checkpoint}; the default local filename is \path{checkpoints/sam3.pt}. The model is wrapped with the official \path{Sam3Processor}, and the response-extraction path is implemented in \path{sam3_runner.py}. The processor converts the image to an 8-bit tensor, resizes it to $1008\times1008$, maps it to floating point, and normalizes each RGB channel using mean and standard deviation $0.5$. The pre-processor evaluation height and width are retained for output restoration.

For an evaluation image, the processor first calls \texttt{set\_image} to create the image state. The prompt \texttt{crack} is encoded by \texttt{forward\_text}, and the resulting text features update \texttt{backbone\_out}. Because no point or box prompt is used, \texttt{\_get\_dummy\_prompt} supplies an empty geometric prompt. The code then calls \texttt{forward\_grounding}. In the specified SAM3 revision, the universal segmentation head returns \texttt{semantic\_seg} together with \texttt{pred\_masks}. For a single image and a single text concept, the extracted semantic tensor is indexed as \mbox{\texttt{semantic\_seg[0,0]}}. Let this logit field be $Z_q$. The response used by SERD is computed as
\begin{equation}
Z_q\xrightarrow{\operatorname{sigmoid}}A_q
\xrightarrow{\text{bilinear restoration}}\widetilde A_q
\xrightarrow{\text{per-image min--max}}R_q.
\end{equation}
No proposal mask, proposal score, connected-component operation, or target-domain annotation enters this response-extraction path.

The native proposal baseline uses the official processor output. Retained proposal masks are restored by bilinear interpolation, sigmoid-activated, binarized at probability 0.5, and merged by union after confidence filtering. Thus, the response and proposal interfaces share the same image backbone, text encoder, grounding computation, and segmentation head but expose different outputs of the same frozen model.

\section{Output-Interface Mismatch Protocol}
\label{app:mismatch_protocol}

The diagnostic in Figure~\ref{fig:interface_mismatch} uses the fixed prompt \texttt{crack} and the same thresholds on every dataset. The active internal-response set is
\begin{equation}
A_i=\{x\in\Omega_i\mid R_{q,i}(x)>0.45\}.
\end{equation}
For native SAM3, the proposal-confidence threshold is $\tau_p=0.4$. Each retained proposal mask is restored and binarized using a sigmoid probability threshold of $0.5$, and $B_i$ is the union of the retained masks. For the ground-truth crack set $P_i$, the mismatch is
\begin{equation}
\mathcal{M}_i=\frac{|(A_i\cap P_i)\setminus B_i|}{|P_i|}.
\end{equation}
The analysis contains 675 Crack500, 157 CamCrack789, 24 CrackMap, 106 DeepCrack, 282 TUT, and 3,390 OmniCrack30k images, for a total of 4,634 crack-bearing images. The reported 8.73\% mismatch, 82.66\% response recall, and 74.66\% proposal recall are equal-weight averages of the six dataset-level image means. Target-test labels are used only for this retrospective diagnostic and do not affect SERD, thresholds, or model parameters.

\section{Segmentation Metrics and SERD-Seg Decoder Details}
\label{app:metric_details}

For Boundary F1, binary boundaries are extracted from predicted and ground-truth masks and expanded using a $5\times5$ elliptical structuring element, corresponding to a matching radius of $\delta=2$ evaluation pixels. Boundary precision and recall are computed by reciprocal matching between the expanded maps, followed by their harmonic mean. For the fixed-radius Boundary IoU, each mask is eroded once with the same $5\times5$ element, the eroded mask is subtracted from the original mask to obtain an internal boundary, and the boundary is dilated once. Boundary IoU is the intersection over union of the resulting predicted and reference boundary bands.

\subsection{Alternative Structural Priors and Optional Binary Readouts}
\label{app:decoder_details}

The structural-prior comparison in Table~\ref{tab5} uses fixed operator settings. Canny is applied to the normalized 8-bit grayscale image with lower and upper thresholds of 50 and 150 and aperture size 3. LoG first applies Gaussian smoothing with $\sigma_x=1.2$, followed by a Laplacian operator; the absolute response is normalized. Morphological black-hat uses a $17\times17$ rectangular structuring element followed by normalization. None of these parameters is tuned per dataset.

For Table~\ref{tab6}, the main branch is $M_{\mathrm{main}}=[R_e\geq0.45]$. The optional weak-response branch is
\begin{equation}
M_W=[R_e\geq0.05]\land[E\geq0.70],\qquad
M_{\mathrm{candidate}}=M_{\mathrm{main}}\lor M_W.
\end{equation}
When $W$ is disabled, $M_{\mathrm{candidate}}=M_{\mathrm{main}}$. Topology filtering defines
\begin{equation}
t_{\mathrm{high}}=\max\!\left(0.30,\operatorname{Percentile}_{99.5}(R_e)\right),
\qquad M_{\mathrm{seed}}=[R_e\geq t_{\mathrm{high}}].
\end{equation}
Eight-connected components are retained only when they intersect $M_{\mathrm{seed}}$ and contain at least 10 pixels. No opening, closing, erosion, or dilation is applied. Thus, the two optional additions introduce five fixed settings: response threshold 0.05, edge threshold 0.70, seed percentile 99.5, seed floor 0.30, and minimum component area 10. They are used only in the ablation and are not part of the default SERD-Seg or SERD-DQ formulation.

\end{document}